\documentclass[preprint,twoside,11pt]{article}

\usepackage{amsthm, amssymb, amsmath}
\usepackage{jmlr2e}
\usepackage{subcaption}   
\usepackage{soul}
\usepackage{tikz}

\usepackage{cleveref}
\usepackage[ruled, vlined]{algorithm2e}
\usepackage{algorithmic}
\usepackage{todonotes}

\tikzset{
    position/.style args={#1:#2 from #3}{
        at=(#3), anchor=#1+180, shift=(#1:#2)
    }
}
    
\DeclareMathOperator*{\E}{\mathbb{E}}
\DeclareMathOperator{\R}{\mathbb{R}}
\DeclareMathOperator*{\Var}{\operatorname{Var}}
\DeclareMathOperator*{\argmin}{arg\,min}
\DeclareMathOperator{\spa}{span}

\jmlrheading{1}{2023}{1-48}{4/00}{10/00}{meila00a}{Lorenz Richter, Leon Sallandt and Nikolas N\"usken}

\ShortHeadings{Tensor trains and robust regression for BSDEs and parabolic PDEs}{Lorenz Richter, Leon Sallandt and Nikolas N\"usken}
\firstpageno{1}

\begin{document}

\title{From continuous-time formulations to \\ discretization schemes: tensor
trains and robust regression \\ for BSDEs and parabolic PDEs}

\author{\name Lorenz Richter \email richter@zib.de \\
       \addr Zuse Institute Berlin, Germany \\
       dida Datenschmiede GmbH, Germany
       \AND
       \name Leon Sallandt \email leon.sallandt@gmail.com \\
       \addr Technische Universit\"at Berlin, Germany
       \AND
       \name Nikolas N\"usken \email nikolas.nusken@kcl.ac.uk \\
       \addr King's College London, UK}

\editor{Some editors}

\maketitle

\begin{abstract}
The numerical approximation of partial differential equations (PDEs) poses formidable challenges in high dimensions since classical grid-based methods suffer from the so-called curse of dimensionality. Recent attempts rely on a combination of Monte Carlo methods and variational formulations, using neural networks for function approximation. Extending previous work \citep{richter2021solving}, we argue that tensor trains provide an appealing framework for parabolic PDEs: The combination of reformulations in terms of backward stochastic differential equations and regression-type methods holds the promise of leveraging latent low-rank structures, enabling both compression and efficient computation. Emphasizing a continuous-time viewpoint, we develop iterative schemes, which differ in terms of computational efficiency and robustness. We demonstrate both theoretically and numerically that our methods can achieve a favorable trade-off between accuracy and computational efficiency.
While previous methods have been either accurate or fast, we have identified a novel numerical strategy that can often combine both of these aspects.
\end{abstract}


\section{Introduction}
Partial differential equations (PDEs) are present in a wide range of scientific and engineering fields. However, when dealing with high-dimensional situations, their numerical handling becomes difficult due to the \emph{curse of dimensionality} that appears in traditional grid-based approaches such as Galerkin methods, finite differences, and others. Recently, however, randomized sampling techniques in combination with powerful function classes and bespoke optimization routines have shown remarkable empirical success \citep{han2018solving,karniadakis2021physics}. Related theoretical results, at least in part confirmatory, are being developed at a fast rate \citep{beck2020overview}, although a complete understanding remains elusive. In principle, stochastic representations as well as variational formulations pertaining to the PDE under consideration are central to those novel approaches, enabling the construction of suitable learning objectives. 

In this work, we build on the conference paper by \cite{richter2021solving}, focussing on semilinear parabolic PDEs and the approximation of their solutions via a well known connection to backward stochastic differential equations (BSDEs), see, for instance \citet{pardoux1998backward}. At the core of this BSDE reformulation is the idea of evaluating (potential) PDE solutions along the paths of a stochastic diffusion process. Therefore, in contrast to more traditional approaches, corresponding numerical methods can be thought of as based on dynamically adaptive random grids, in principle holding the promise of scaling according to dimension-free Monte Carlo approximation rates. 

Solving BSDEs requires the choice of both an appropriate function class (ideally incorporating sparsity and compression features for functions defined on high-dimensional spaces) as well as of an efficient time stepping scheme. As argued in \citep{richter2021solving}, the tensor train format \citep{oseledets2011tensor} provides an appealing framework that addresses both desiderata. Indeed, tensor train representations ensure good 
scalability to high-dimensional settings by relying on constrained combinations of functions of a single real variable.
In the presence of low-dimensional latent structures (and favorable alignment of those with the chosen tensor train representation), this construction alleviates the curse of dimensionality. At the same time, the specific make-up of tensor trains allows for iterative least-squares updates to efficiently compute the solutions to regression-type problems typical of discrete-time schemes for BSDEs \citep{bouchard2004discrete,gobet2005regression}. 
\\

\textbf{Conditional expectations and robust regression.} The numerical treatment of BSDEs is intimately related to the computation of conditional expectations; many current methods therefore rely on techniques borrowed from statistics, and in particular from regression analysis \citep[Section 4]{chessari2021numerical}. In this introductory paragraph, however, we would like to argue that the settings in statistics and BSDEs differ in a subtle way, and that recognizing and leveraging these differences can lead to improved schemes for BSDEs which will be detailed in the upcoming sections.
To convey the main idea in an abstract context, let us assume that the real-valued random variables $X$ and $Y$ have finite second moments and are connected through the relation
\begin{equation}
\label{eq:regression}
Y = f^*(X) + \varepsilon,   
\end{equation}
with a noise variable $\varepsilon$ that has finite variance and satisfies $\mathbb{E}[\varepsilon|X] = 0$. Under mild conditions on $f^*$, the task of recovering it from $K$ observations $(X_k,Y_k)_{k=1}^K$ (that is, from $K$ realizations of the joint variable $(X,Y)$) can be approached using the relation 
\begin{equation}
\label{eq:conditional expectations}
f^*(\cdot) = \mathbb{E} [Y | X= \cdot] = \argmin_{{f}} \mathbb{E} \left[ \left(Y - {f}(X)\right)^2\right],
\end{equation}
which directly follows from \eqref{eq:regression} and the characterization of conditional expectations in terms of $L^2$-projections \cite[Corollary 8.17]{klenke2013probability}. Indeed, given a parameterization $f_{\theta}$ of candidate approximations for $f^*$, equation \eqref{eq:conditional expectations} motivates setting $f^* \approx f_{\theta^*}$ with $\theta^* \in \argmin \mathcal{L}$, where  
the loss function $\mathcal{L}$ is given by  
\begin{equation}
\label{eq:regression objective}
\mathcal{L}^{(K)}(\theta) = \frac{1}{K} \sum_{k=1}^K \left(Y_k - f_\theta(X_k)  \right)^2.
\end{equation}
Notably, the least squares objective \eqref{eq:regression objective} can be evaluated without access to the noise realizations $(\varepsilon_k)_{k=1}^K$ and those are indeed typically unavailable in classical statistical settings where \eqref{eq:regression} could for instance model a phenomenon found in nature. In the context of BSDEs, on the other hand, the perturbations $(\varepsilon_k)_{k=1}^K$ are generated within the algorithm and may thus be used in the formulation of optimization objectives. In particular, we can modify the loss \eqref{eq:regression objective} as follows,
\begin{equation}
\label{eq:robust loss}
\mathcal{L}^{(K)}_{\mathrm{robust}}(\theta) = \frac{1}{K} \sum_{k=1}^K \left( Y_k - f_\theta(X_k) - \varepsilon_k \right)^2, 
\end{equation}
directly enforcing the relation \eqref{eq:regression} on the basis of the samples $(X_k,Y_k,\varepsilon_k)_{k=1}^K$. As alluded to in the notation, we expect the inclusion of $(\varepsilon_k)_{k=1}^K$ to have a variance-reducing effect that can make the numerical procedure more robust. Indeed, it is straightforward to verify that $\partial_{\theta} \mathcal{L}^{(K)}_{\mathrm{robust}}|_{f_\theta = f^*} = 0$ \emph{almost surely}, that is, the gradient of the objective \eqref{eq:robust loss} vanishes at the optimum \emph{notwithstanding the fact that a finite-sample approximation is used}. In contrast, we see that $\partial_\theta \mathcal{L}^{(K)}|_{f_\theta = f^*}$ is random with mean zero,  $\mathbb{E}[\partial_\theta \mathcal{L}^{(K)}|_{f = f_\theta}] = 0$, that is, the objective \eqref{eq:regression objective} requires the law of large number limit $K \rightarrow \infty$ in order to reliably identify the optimizer $f^*$. As a consequence, procedures based on \eqref{eq:regression objective} may become unstable in the regime $f_\theta \approx f^*$ due to a low signal-to-noise ratio. 
\\

\textbf{Contributions.} 
This paper builds on work by \cite{richter2021solving}, and we develop the arguments from the previous paragraphs in the context of BSDEs, thus providing a comprehensive analysis on related numerical stability issues (see in particular \eqref{eq: residual loss} and \eqref{eq: projection loss} below, and compare with \eqref{eq:robust loss} and \eqref{eq:regression objective}, respectively). Whilst our exposition in Section \ref{sec: BSDE} follows in large parts the existing vast literature on the numerical treatment of BSDEs (see, for instance, \cite{chessari2021numerical} for a survey), we place particular emphasis on formulations in continuous time, turning to time-discretizations at the last step of the derivation, see Section \ref{sec:dis time}. This approach allows us to effortlessly construct explicit and implicit time stepping schemes in combination with losses of the form \eqref{eq:regression objective} and \eqref{eq:robust loss},
extending the methodology put forward in \citep{richter2021solving} and leading in particular to the robust explicit loss in \eqref{eq: discrete L_BSDE_exp}, that so far appears to have attracted little attention. We develop a tensor train based scheme for its optimisation (see Section \ref{sec: handling L_BSDE_exp}) and numerically investigate its performance:
\\

\textbf{Numerical evaluation.}
Explicit and implicit numerical schemes potentially lead to a trade-off between computational cost and stability (see  \citep{chassagneux2015numerical} for a related discussion in a slightly different context). In our numerical experiments, however, we do not observe significant improvements in accuracy or stability imparted by the implicit schemes, whereas the numerical overhead is substantial.
We thus conjecture that for high dimensional problems, which are the focus of this paper, sampling errors exceed discretization errors.

The BSDE-versions of the robust loss \eqref{eq:robust loss} overall significantly outperform methods based on \eqref{eq:regression objective}
in terms of approximation accuracy.
Going beyond the abstract setting from \eqref{eq:regression}, we observe in one experiment that concrete implementations of \eqref{eq:regression objective} and \eqref{eq:robust loss} in the BSDE setting lead to schemes that tend to shift emphasis from the accuracy of the PDE solution to its gradient (see Remark \ref{rem:gradient forcing} below and the experiments in Section \ref{sec:HJB doublewell}). This phenomenon (at the moment supported only by preliminary numerical evidence) might be of particular interest in the context of stochastic optimal control. 

Combining the insights from these observations, we would like to advertise the explicit and robust loss defined in \eqref{eq: discrete L_BSDE_exp}, which, to the best of our knowledge, has so far not been used in numerical experiments. We believe that a more in-depth analysis of its properties is a promising avenue for future work.

\subsection{Setting and notation}
\label{sec:setting}

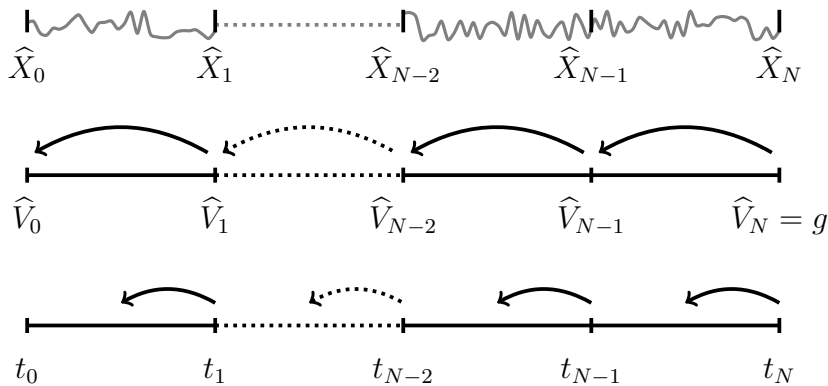
\begin{figure}[h]
    \centering
\begin{tikzpicture}[%
    every node/.style={
        font=\scriptsize,
        text height=1ex,
        text depth=.25ex,
    },
]

\pgfmathsetseed{1}
\draw[very thick, gray, smooth, domain=0:25] plot[samples = 30] (\x / 10, 0.2 * rand);
\draw[dotted, gray, line width = 0.5mm] (2.5,0) -- (5,0);
\draw[very thick, gray, smooth, domain=50:100] plot[samples = 60] (\x / 10, 0.2 * rand);
\foreach \x in {0,2.5,...,10}{
    \draw[black, line width = 0.5mm] (\x cm, 0.2) -- (\x cm, -0.1);
}

\node[anchor=north] at (0,-0.4) {\large$\widehat{X}_0$};
\node[anchor=north] at (2.5,-0.4) {\large$\widehat{X}_1$};
\node[anchor=north] at (5,-0.4) {\large$\widehat{X}_{N-2}$};
\node[anchor=north] at (7.5,-0.4) {\large$\widehat{X}_{N-1}$};
\node[anchor=north] at (10,-0.4) {\large$\widehat{X}_N$};

\draw[black, line width = 0.5mm] (0,-2) -- (2.5,-2);
\draw[dotted, black, line width = 0.5mm] (2.5,-2) -- (5,-2);
\draw[black, line width = 0.5mm] (5,-2) -- (7.5,-2);
\draw[black, line width = 0.5mm] (7.5,-2) -- (10,-2);

\foreach \x in {0,2.5,...,10}{
    \draw[black, line width = 0.5mm] (\x cm, -1.9) -- (\x cm, -2.1);
}

\draw[<-, black, line width = 0.5mm] (0.1,-1.7) to[bend left] (2.4,-1.7);
\draw[<-, black, line width = 0.5mm, dotted] (2.6,-1.7) to[bend left] (4.9,-1.7);
\draw[<-, black, line width = 0.5mm] (5.1,-1.7) to[bend left] (7.4,-1.7);
\draw[<-, black, line width = 0.5mm] (7.6,-1.7) to[bend left] (9.9,-1.7);

\node[anchor=north] at (0,-2.4) {\large$\widehat{V}_0$};
\node[anchor=north] at (2.5,-2.4) {\large$\widehat{V}_1$};
\node[anchor=north] at (5,-2.4) {\large$\widehat{V}_{N-2}$};
\node[anchor=north] at (7.5,-2.4) {\large$\widehat{V}_{N-1}$};
\node[anchor=north] at (10,-2.4) {\large$\widehat{V}_N = g$};


\draw[black, line width = 0.5mm] (0,-4) -- (2.5,-4);
\draw[dotted, black, line width = 0.5mm] (2.5,-4) -- (5,-4);
\draw[black, line width = 0.5mm] (5,-4) -- (7.5,-4);
\draw[black, line width = 0.5mm] (7.5,-4) -- (10,-4);

\foreach \x in {0,2.5,...,10}{
    \draw[black, line width = 0.5mm] (\x cm, -3.9) -- (\x cm, -4.1);
}

\draw[<-, black, line width = 0.5mm] (1.25,-3.7) to[bend left] (2.5,-3.7);
\draw[<-, black, line width = 0.5mm, dotted] (3.75,-3.7) to[bend left] (5.0,-3.7);
\draw[<-, black, line width = 0.5mm] (6.25,-3.7) to[bend left] (7.5,-3.7); 
\draw[<-, black, line width = 0.5mm] (8.75,-3.7) to[bend left] (10.0,-3.7);

\node[anchor=north] at (0,-4.4) {\large$t_0$};
\node[anchor=north] at (2.5,-4.4) {\large$t_1$};
\node[anchor=north] at (5,-4.4) {\large$t_{N-2}$};
\node[anchor=north] at (7.5,-4.4) {\large$t_{N-1}$};
\node[anchor=north] at (10,-4.4) {\large$t_N$};

\end{tikzpicture}
\caption{Schematic version of backward iterations for the numerical approximation of the solution to the PDE \eqref{eq: definition general PDE}. The procedure starts at the known terminal value $V(\cdot, T) =  g$ and gradually iterates backward in time along a time grid $0 = t_0 < t_1 < \cdots < t_N = T$ and along trajectories of the (discretized) forward process $\widehat{X}$.}
\label{fig: backward iteration scheme}
\end{figure}

In this paper, we focus on semi-linear parabolic PDEs, which have the following form:
\begin{subequations}
\label{eq: definition general PDE}
\begin{align}
\label{eq:PDE}
    (\partial_t + L) V(x, t) + h(x, t, V(x, t), (\sigma^\top \nabla V)(x, t)) &= 0, & (x, t) \in {\R}^d \times [0, T), \\
    \label{eq:terminal pde}
    V(x, T) &= g(x), & x \in \R^d,
\end{align}
\end{subequations}
where $h \in C(\R^d \times [0, T] \times \R \times \R^d, \R) $ specifies the nonlinearity, $g \in C(\R^d, \R)$ is the terminal condition, and
\begin{equation}
\label{eq: infinitesimal generator}
    L = \frac{1}{2} \sum_{i,j=1}^d (\sigma \sigma^\top)_{ij}(x,t) \partial_{x_i} \partial_{x_j} + \sum_{i=1}^d b_i(x,t) \partial_{x_i}
\end{equation}
is a second-order (elliptic) diffential operator containing the coefficients $b \in C(\R^d \times [0, T], \R^d)$ and $\sigma \in C(\R^d \times [0, T], \R^{d\times d})$. We assume that the matrix $\sigma \sigma^\top(x)$ is nondegenerate for all $x \in \mathbb{R}^d$, and that \eqref{eq: definition general PDE} admits a unique (classical) solution $V:\mathbb{R}^d \times [0,T] \rightarrow \mathbb{R}$. The methods considered in this paper can broadly be summarized as follows:

\begin{enumerate}
    \item For every instance $t_n$ within a time grid  $0 = t_0 < t_1 < \cdots < t_N = T$, we produce `spatial grid points' $(\widehat{X}^{(k)}_{n})_{k=1}^K$ on the basis of which the solution $V$ shall be approximated. More precisely, we will aim to construct $\widehat{V}_n:\mathbb{R}^d \rightarrow \mathbb{R}$ such that
    \begin{equation}
    \label{eq:approx intro}
    \widehat{V}_n(\widehat{X}_n^k) \approx V(\widehat{X}_n^k,t_n), \qquad \text{for all } n = 1,\ldots,N, \quad k = 1,\ldots,K.  
    \end{equation}
    Crucially, the grid points $(\widehat{X}^{(k)}_{n})_{k=1}^K$ will be (approximate) samples from the diffusion process
    \begin{equation}
    \label{eq:diffusion intro}
    \mathrm dX_s = b(X_s, s) \, \mathrm ds + \sigma(X_s, s) \, \mathrm d W_s
    \end{equation}    
    associated to the operator $L$ defined in \eqref{eq: infinitesimal generator}. The connection between \eqref{eq: infinitesimal generator} and \eqref{eq:diffusion intro} underlying the generation of $(\widehat{X}^{(k)}_{n})_{k=1}^K$ is at the heart of the BSDE approach, see Section \ref{sec: BSDE}. Since $V(\cdot,T)=g$ is known from the terminal condition \eqref{eq:terminal pde}, we will address \eqref{eq:approx intro} iteratively backwards in time using the updates $\widehat{V}_n \rightsquigarrow \widehat{V}_{n-1}$, see Figure \ref{fig: backward iteration scheme}. This procedure corresponds to the backward process in the BSDE, see \Cref{sec: BSDE}.
    
    \item To achieve the approximation \eqref{eq:approx intro}, we reformulate this equation as a least squares regression problem (utilizing the approximation previously obtained at $t_{n+1}$), in the spirit of \eqref{eq:regression objective} and \eqref{eq:robust loss}. For $\widehat{V}_n$, we use a tensor train ansatz of the form
    \begin{equation}
    \widehat{V}_n(x_1,\ldots,x_d) = \sum a_{i_1,\ldots,i_d} \phi_{i_1}(x_1) \cdot \ldots \cdot \phi_{i_d}(x_d),
    \end{equation}
    with fixed ansatz functions $\phi_i:\mathbb{R} \rightarrow \mathbb{R}$ that importantly only take real numbers (of dimension one) as inputs, hence could be thought of as one-dimensional building blocks. To ensure scalability to the high-dimensional setting, the tensor train approach places stringent low-rank conditions on the coefficients  $a_{i_1,\ldots,i_d}$, see Section \ref{sec: BSDEs via TTs}.
\end{enumerate}

\begin{remark}[Generalizations]
The discussion in this paper can be generalized in multiple ways beyond the semi-linear parabolic PDE \eqref{eq: definition general PDE}. First, the terminal value problem can be converted into an initial value problem by applying the time inversion $t \mapsto T - t$. Furthermore, generalizations to elliptic PDEs as well as to PDEs on bounded domains are in principle possible (cf. \cite{nusken2021interpolating}). However, the approaches considered in this paper based on backward iterations would need nontrivial adaptations due the change in boundary conditions. Finally, one can also aim at targeting fully nonlinear PDEs, see e.g. \cite{beck2019machine} and \cite{pham2021neural}.
\end{remark}

\subsection{Previous work}

BSDEs have been introduced in \citet{bismut1973conjugate} and were studied systematically in \citet{pardoux1990adapted}. We refer the reader to \citet{pardoux1998backward} for an overview that highlights the connections of BSDEs to elliptic and parabolic PDEs. A recent survey about numerical methods for BSDEs can be found in \citet{chessari2021numerical}, tracing the development from regression based approaches in finance \citep{longstaff2001valuing} through the works \cite{ma2002numberical,bouchard2004discrete,zhang2004numerical,gobet2005regression, gobet2007error} towards recent developments involving deep learning \citep{beck2019deep,pham2021neural}. The present article is an extension of the conference paper \cite{richter2021solving}, laying particular emphasis on robustness and further analyzing the trade-offs between explicit and implicit discretization schemes. For \cite{richter2021solving} and this paper, the following aspects are particularly relevant:
\\

\emph{Function classes.} Tensor trains were first introduced to the mathematical community in  \cite{oseledets2011tensor}, with their foundations based in quantum physics and known as `matrix product states'. They can be seen as a special case of hierarchical tensor networks, which have been developed in \cite{Hackbusch-2010}. To obtain comprehensive surveys and further information on tensor trains, we recommend referring to the following sources: \cite{Hackbusch-Acta, Hackbusch2014, Legeza-Schneider, Bachmayr-Uschmajew-Schneider}. \\

\emph{PDE solvers.} The numerical treatment of PDEs has seen recent advances via the idea of combining sophisticated function approximation with Monte Carlo sampling. Relying on deep learning techniques, we highlight \cite{weinan2017deep}, which builds on a variational formulation based on BSDEs, see also \cite{nusken2021interpolating}. \cite{beck2019deep} and \cite{hure2020deep} combine deep learning attempts with backwards schemes based on Feynman-Kac and BSDEs, respectively. For recent overviews of deep learning for the approximation of soutions to PDEs we refer to \cite{e2020algorithms} and \cite{richter2021phd}. Tensor trains have been used to approximate PDE solutions as well. For the treatment of parametric PDEs we refer to \cite{dolgov2015polynomial, eigel2017adaptive, dektor2020rank}. The approximation of Hamilton-Jacobi-Bellman PDEs with tensor trains can be found in \cite{horowitz2014linear, stefansson2016sequential, gorodetsky2018high,dolgov2019tensor,oster2019approximating,fackeldey2020approximative, chen2021tensor} and we refer to \cite{khoromskij2012tensors, kormann2015semi, lubasch2018multigrid} for further types of PDEs.  \\

\emph{Robustness.} The idea of incorporating the It\^{o} integral into the objective dates back to \cite{gobet2005regression}, where fixed point iterations are used for solving BSDEs. Using related robust and variance-reduced objectives for gradient descent based algorithms has been employed in \cite{zhou2021actor} for Hamilton-Jacobi-Bellman equations and in \cite{richter2022robust} for linear PDEs. For an extensive analysis of the robustness of losses and their gradients we refer to \cite{nusken2020solving}.
\\

\emph{Discretizations.} In algorithmic contexts, time-continuous BSDEs have to be discretized. \cite{chassagneux2014runge} study Runge-Kutta schemes for the approximation of the deterministic integral in order to potentially improve the convergence rate. Linear multistep methods are considered in \cite{chassagneux2014linear} and the numerical implications of explicit vs. implicit schemes are investigated in \cite{chassagneux2015numerical}. For a numerical analysis of the deep BSDE algorithm we refer to \cite{han2020convergence}.

\subsection{Outline of the article} The paper is structured as follows. In \Cref{sec: BSDE} we give an introduction to BSDEs and highlight their interpretation as a stochastic representation of the  PDE \eqref{eq: definition general PDE}, deriving appropriate loss functionals. In particular,  \Cref{sec: numerical approximation of BSDEs} focuses on continuous-time formulations, while  numerical aspects that become apparent in the discretization of BSDEs will be discussed in \Cref{sec:dis time}. Thereafter, \Cref{sec: robustness properties} investigates the identified losses with respect to statistical robustness properties. \Cref{sec: BSDEs via TTs} is then devoted to the tensor train format as a valid means for function approximation, encompassing complexity (\Cref{sec:common ops}), optimization (\Cref{sec:opt TT manifold}) and gradient-dependent loss functionals (\Cref{sec: handling L_BSDE_exp}). Finally, in \Cref{sec: numerical examples} we illustrate our theoretical findings on multiple numerical examples, before \Cref{sec: conclusion} ends with a conclusion and outlook.

\section{Solving PDEs via BSDEs}
\label{sec: BSDE}

In this section we review the connection between semi-linear PDEs of the type \eqref{eq: definition general PDE} and backward stochastic differential equations (BSDEs), going back to  \citet{bismut1973conjugate} and \citet{pardoux1990adapted}.
Roughly, BSDEs can be linked to the Feynman-Kac formula \citep[Chapter 8]{oksendal2013stochastic}, offering stochastic representations of solutions to PDEs (\cite{pardoux1998backward}), however also in the nonlinear setting, such as the one stated in \eqref{eq: definition general PDE}. The  connection between PDEs and BSDEs has been studied extensively in the last decades (see e.g. \cite{bouchard2009discrete}, \cite{fahim2011probabilistic} and \cite{weinan2017deep}) and we refer to \cite{pham2009continuous}, \cite{touzi2012optimal} and \cite{zhang2017backward} for excellent introductions to the subject.

In a nutshell, BSDEs can be derived from evaluating the solution of the PDE \eqref{eq: definition general PDE} along the stochastic process \eqref{eq:diffusion intro}, here restated for convenience,
\begin{equation}
\label{eq: fordward SDE}
    \mathrm dX_s = b(X_s, s) \, \mathrm ds + \sigma(X_s, s) \, \mathrm d W_s, \quad X_0 = x_0,
\end{equation}
where $b \in C(\R^d \times [0, T], \R^d)$ and $\sigma \in C(\R^d \times [0, T], \R^{d\times d})$ refer to the same functions as in \eqref{eq: infinitesimal generator}.
Conceptually speaking, the SDE \eqref{eq: fordward SDE} provides a (stochastic) grid on which the PDE solution is obtained, not unlike the equidistant (deterministic) grids used in finite-difference schemes \citep{johnson2012numerical}.
To spell this out, let us assume that $V \in C^{2,1}(\R^d \times [0,T], \R)$ is a classical solution to the PDE \eqref{eq: definition general PDE} and define the process 
\begin{equation}
\label{eq: def Y}
    Y_s = V(X_s, s),
\end{equation}
evaluating $V$ along $(X_s)_{0 \le s \le T}$
on the time interval $s \in [0, T]$. An application of It\^{o}'s formula shows that
\begin{equation}
    V(X_T, T) - V(X_0, 0) = \int_0^T (\partial_t + L) V(X_s, s) \,\mathrm ds + \int_0^T (\sigma^\top \nabla V)(X_s, s)\cdot \mathrm dW_s,
\end{equation}
and therefore, using \eqref{eq: definition general PDE},
\begin{equation}
\label{eq:BSDE integral}
    g(X_T) = V(X_0, 0) - \int_0^T h(X_s, s, V(X_s, s), (\sigma^\top \nabla V)(X_s, s)) \,\mathrm ds + \int_0^T (\sigma^\top \nabla V)(X_s, s)\cdot \mathrm dW_s.
\end{equation}
Further introducing the shorthand notation 
\begin{equation}
\label{eq: def Z}
Z_s = (\sigma^\top \nabla V)(X_s, s),
\end{equation}
we can equivalently write \eqref{eq:BSDE integral} as
\begin{equation}
\label{eq: BSDE}
    \mathrm dY_s = -h(X_s, s, Y_s, Z_s) \,\mathrm ds + Z_s \cdot \mathrm dW_s, \qquad Y_T = g(X_T).
\end{equation}
Solutions to the equations \eqref{eq: fordward SDE} and \eqref{eq: BSDE} are triplets $(X, Y, Z)$, where $X$ is called the forward process, whilst $Y$ and $Z$ are referred to as backward processes, owing to the terminal condition in \eqref{eq: BSDE}. However, $(X_s)_{s \in [0,T]}$, $(Y_s)_{s \in [0,T]}$ and $(Z_s)_{s \in [0,T]}$  are adapted to the filtration $(\mathcal{F}_s)_{s \in [0,T]}$ generated by the Brownian motion $(W_s)_{s \in [0,T]}$, and therefore, in a causal sense, should be thought of as evolving forward in time.
Together with this adaptedness condition, the equations \eqref{eq: fordward SDE} and \eqref{eq: BSDE} provide three constraints for the triplet $(X,Y,Z)$, hence existence and uniqueness under appropriate conditions is plausible. For rigorous results we refer to \citet{el1997backward}, \citet[Theorem 10.2]{touzi2012optimal}, \citet[Theorem 4.3.1, Theorem 7.3.3]{zhang2017backward} and \citet{kobylanski2000backward}.

\begin{remark}[Controlled forward processes]
Using the same arguments, we can generalize the system of BSDEs specified in \eqref{eq: fordward SDE} and \eqref{eq: BSDE} by adding a control term $v \in C(\R^d \times [0, T], \R^d)$ to the forward process in \eqref{eq: fordward SDE}. Ensuring $Y_s^v = V(X_s^v, s)$ and $Z_s^v = (\sigma^\top\nabla V)(X_s^v, s)$ leads to a compensation term in the backward process,
\begin{subequations}
\begin{align}
    \mathrm d X^v_s &= \left(b(X^v_s, s) + \sigma(X_s^v, s) v(X_s^v, s)\right) \mathrm ds + \sigma(X^v_s, s) \, \mathrm dW_s, \qquad & X^v_0 = x_\mathrm{0}, \\
    \mathrm d Y^v_s &= \left(-h(X^v_s, s, Y^v_s, Z^v_s) + v(X_s^v, s) \cdot Z_s^v \right)  \mathrm ds +  Z^v_s \cdot \mathrm dW_s, \qquad & Y^v_T = g(X_T^v).
\end{align}
\end{subequations}
The control $v$ in the forward process may be beneficial from the computational point of view, pushing the process into regions of interest or reducing the variance of Monte Carlo estimators, cf. \citet{hartmann2019variational}.
\end{remark}

\subsection{Solving BSDEs: loss functionals in continuous time}
\label{sec: numerical approximation of BSDEs}

The BSDE formulation \eqref{eq: BSDE} opens the door for Monte Carlo approaches towards solutions to the PDE \eqref{eq: definition general PDE} by solving for the triplet $(X,Y,Z)$ and using the correspondences \eqref{eq: def Y} and \eqref{eq: def Z}. Numerical approaches can broadly be classified into two branches: 
One can either approximate solutions to the PDE \eqref{eq: definition general PDE} on the entire time interval $[0, T]$ at once, directly incorporating the time dependence (\cite{weinan2017deep}, \cite{raissi2018forward}, \cite{nusken2020solving}). Alternatively, one can divide the problem into multiple subproblems and approach those one after another (\cite{bouchard2004discrete}, \cite{gobet2005regression}, \cite{hure2020deep}). In this work we focus on the latter attempt.

In the spirit of the dynamic programming principle from optimal control theory \citep{fleming2012deterministic}, the idea is to solve for the backward process on separate disjoint time intervals defined by the discretization
\begin{equation}
\label{eq:time grid}
    0 = t_0 < t_ 1 < \cdots < t_N = T,
\end{equation} 
thereby dividing the problem into a sequence of subproblems.
The algorithm proceeds backwards in time, starting with the last interval $[t_{N-1},t_N]$ and using the terminal condition $Y_T = V(X_T, T) = g(X_T)$. Subsequently, the computations on each interval will rely on approximations from previously approached intervals, with the solution at time $t_n$ providing the terminal condition for the subproblem posed on the interval $[t_{n-1},t_n]$, for $n=1,\ldots,N-1$. As an illustration, we refer to Figure \ref{fig: backward iteration scheme}. 

To approach the subproblem on the time interval $[t_{n}, t_{n+1}]$, we first
state the backward process \eqref{eq: BSDE} in an integrated version as
\begin{equation}
         Y_{t_{n+1}} - Y_{t_n} = -\int_{t_n}^{t_{n+1}} h(X_s, s, Y_s, Z_s) \,\mathrm ds + \int_{t_n}^{t_{n+1}} Z_s \cdot \mathrm dW_s,
\end{equation}
or, equivalently, as
\begin{align}
\label{eq: Ito t_n t_n+1}
\begin{split}
         V(X_{t_{n+1}}, t_{n+1}) - V(X_{t_{n}}, t_n) &= -\int_{t_n}^{t_{n+1}} h(X_s, s, V(X_s, s), (\sigma^\top \nabla V)(X_s, s)) \,\mathrm ds \\
         &\qquad\qquad\qquad\qquad\qquad + \int_{t_n}^{t_{n+1}} (\sigma^\top \nabla V)(X_s, s) \cdot \mathrm dW_s,
\end{split}
\end{align}
both of which hold true almost surely. According to \citet[Chapter 6]{pham2009continuous} and slightly simplifying, the formulation \eqref{eq: Ito t_n t_n+1} is equivalent to the PDE \eqref{eq:PDE} on $[t_n,t_{n+1}]$, in the sense that \eqref{eq: Ito t_n t_n+1} holds almost surely if and only if $V$ satisfies \eqref{eq:PDE}.  

Based on this observation, we may now aim to construct suitable loss functionals
\begin{equation}
\label{eq: general loss}
    \mathcal{L}_n:\mathcal{H} \to \R_{\ge 0}
\end{equation}
that specify the misfit between an approximating function $\varphi$ and the solution $V$ on the interval $[t_n, t_{n+1}]$. Here, $\mathcal{H}$ is an appropriate function class to be chosen later on, for now assumed to contain the solution, i.e. $V \in \mathcal{H}$. More precisely, the loss \eqref{eq: general loss} shall be minimal if and only if the approximation is exact, i.e.
\begin{equation}
\label{eq: abstract loss property}
    \varphi^* \in \argmin_{\varphi \in \mathcal{H}} \mathcal{L}_n(\varphi)  \qquad \Longleftrightarrow \qquad  \varphi^*(x, t)  = V(x, t) \quad \forall\, (x, t) \in \R^d \times [t_n, t_{n+1}].
\end{equation}
In line with the iterative procedure depicted in Figure \ref{fig: backward iteration scheme}, $\mathcal{L}_n$ will typically be constructed using the approximation $\varphi$ obtained in the succeeding time interval $[t_{n+1},t_{n+2}]$; in this situation we require \eqref{eq: abstract loss property} to hold provided that this approximation is exact, i.e. $\varphi = V$ on $[t_{n+1},t_{n+2}]$.\footnote{In fact, as the methods proceed locally in time, we will only require $\varphi(\cdot,t_{n+1}) = V(\cdot,t_{n+1})$.}

In the following we review two loss functionals that are based on \eqref{eq: Ito t_n t_n+1} and satisfy \eqref{eq: abstract loss property}. 
The first loss,
\begin{align}
\label{eq: residual loss}
    \begin{split}
    {\mathcal{L}}_\text{BSDE}(\varphi) &= \E\Bigg[\Bigg(\varphi(X_{t_n}, t_n) - V(X_{t_{n+1}}, t_{n+1}) - \int_{t_n}^{t_{n+1}} h(X_s, s, \varphi(X_s, s), (\sigma^\top \nabla \varphi)(X_s, s))\,\mathrm ds \\
    &\qquad\qquad\qquad\qquad\qquad\qquad\qquad\qquad + \int_{t_n}^{t_{n+1}} (\sigma^\top \nabla \varphi)(X_s, s)\cdot \mathrm dW_s \Bigg)^2 \Bigg],
    \end{split}
\end{align}
measures the misfit in the BSDE \eqref{eq: Ito t_n t_n+1}
when the solution $V$ is replaced by the approximating function $\varphi$, in mean square sense. The formulation in \eqref{eq: residual loss} is in the spirit of \cite{gobet2005regression}, see equation (4) within in this work. As alluded to above, the loss $\mathcal{L}_{\mathrm{BSDE}}$ assumes knowledge of $V(X_{t_{n+1}}, t_{n+1})$, which can be viewed as a terminal condition for the solution on the time interval $[t_n, t_{n+1}]$.
In the context of the iterative scheme from Figure \ref{fig: backward iteration scheme}, we will set $V(X_{t_{n+1}},t_{n+1}) \approx \varphi(X_{t_{n+1}},t_{n+1})$, using the approximation obtained previously in the succeding time interval (cf. 
 Remark \ref{rem: error propagation}).
Note that here and in the following  we omit the index $n$  indicating the time interval in $\mathcal{L}_{\mathrm{BSDE}}$ for notational convenience. \par\bigskip

An alternative loss can be derived by applying the conditional expectation with respect to the Brownian filtration $\mathcal{F}_{t_n} = \sigma(W_s: \,\, 0 \le s \le t_n)$ to \eqref{eq: Ito t_n t_n+1}, yielding
\begin{subequations}
\label{eq: conditional expectation}
\begin{align}
\nonumber
         & V(X_{t_{n}}, t_n)  = \E\left[  V(X_{t_{n+1}}, t_{n+1}) + \int_{t_n}^{t_{n+1}} h(X_s, s, V(X_s, s), (\sigma^\top \nabla V)(X_s, s))\,\mathrm ds \Bigg| \mathcal{F}_{t_n}\right]
         \quad \,\, (\ref*{eq: conditional expectation})
         \\
        & = \argmin_{\varphi} \E\Bigg[\Big(\varphi(X_{t_n}, t_n) - V(X_{t_{n+1}}, t_{n+1}) 
         - \int_{t_n}^{t_{n+1}} h(X_s, s, V(X_s, s), (\sigma^\top \nabla V)(X_s, s))\,\mathrm ds\Big)^2 \Bigg],
         \nonumber
\end{align}
\end{subequations}
where in the first line we have used the martingale property of the (It\^o) stochastic integral as well as the fact that $V(X_{t_n},t_n)$ is $\mathcal{F}_{t_n}$-measurable. In the second line, we have reformulated the conditional expectation in terms of a least-squares minimization (or projection), see \citet[Corollary 8.17]{klenke2013probability}. 
The relation \eqref{eq: projection loss} straightforwardly suggests the projection loss
\begin{align}
\label{eq: projection loss}
    {\mathcal{L}}_{\text{proj}}(\varphi) &=  \E\Bigg[\Big(\varphi(X_{t_n}, t_n) - V(X_{t_{n+1}}, t_{n+1})
    \\
    & \qquad \qquad - \int_{t_n}^{t_{n+1}} h(X_s, s, V(X_s, s), (\sigma^\top \nabla V)(X_s, s))\,\mathrm ds\Big)^2 \Bigg],
    \nonumber
\end{align}
see \citet[Section 3.1]{chessari2021numerical} and references therein.

Comparing  ${\mathcal{L}}_{\text{BSDE}}$ to $\mathcal{L}_{\mathrm{proj}}$, we see that the It\^{o} integral has been removed, essentially by applying the conditional expectation in \eqref{eq: Ito t_n t_n+1}. Conversely, the passage from $\mathcal{L}_{\textrm{proj}}$ to $\mathcal{L}_{\textrm{BSDE}}$ has an interpretation in terms of control variates, see \cite{robert2004monte} for a general introduction and e.g. \cite{zhou2021actor} or \cite{richter2022robust} for applications of control variates to linear PDEs. Indeed, starting with \eqref{eq: conditional expectation} and using the fact that the stochastic integral in the second line of \eqref{eq: residual loss} has (conditional) expectation zero, we may add this term to \eqref{eq: conditional expectation} in the hope of reducing the variance of corresponding Monte Carlo estimators. At this point, we would also like to point out the conceptual similarity between \eqref{eq: projection loss} and \eqref{eq:regression} on the one hand, and \eqref{eq: residual loss} and \eqref{eq:robust loss} on the other hand.

Further, note that $\mathcal{L}_{\mathrm{proj}}$ contains the solution $V$ in the Riemann integral (in contrast to the corresponding term in $\mathcal{L}_{\mathrm{BSDE}}$, which depends on $\varphi$). For discretization schemes, this implies that only an approximation of the integral by its right end point contribution will lead to feasible numerical strategies, making use of $V(X_{t_{n+1}}, t_{n+1}) \approx \varphi(X_{t_{n+1}}, t_{n+1})$ obtained in the previous iteration on the interval $[t_{n+1}, t_{n+2}]$. As we will see in Section \ref{sec:dis time} below, the loss $\mathcal{L}_{\mathrm{BSDE}}$ allows for more flexible discretizations in time. 

\begin{remark}[(Non-)vanishing losses]
\label{rem: L_BSDE to zero}
Substituting $\varphi = V$ into $\mathcal{L}_{\mathrm{proj}}$ and $\mathcal{L}_{\mathrm{BSDE}}$ as well as using the relation \eqref{eq: Ito t_n t_n+1} we see that
\begin{equation}
\label{eq:vanishing}
   \min_{\varphi \in \mathcal{H}} {\mathcal{L}}_{\mathrm{proj}}(\varphi) = \E\left[\int_{t_n}^{t_{n+1}} |(\sigma^\top \nabla V)(X_s, s)|^2\,\mathrm{d}s\right], \qquad\qquad    \min_{\varphi \in \mathcal{H}} {\mathcal{L}}_{\mathrm{BSDE}}(\varphi) = 0.
\end{equation}
Therefore, the accuracy of an approximation $\varphi \in \mathcal{H}$ may be monitored in an online-fashion in algorithms that are based on $\mathcal{L}_{\mathrm{BSDE}}$. For methods based on $\mathcal{L}_{\mathrm{proj}}$, the same may prove challenging since the expectation in \eqref{eq:vanishing} will rarely be available in closed form.
\end{remark}

\subsection{Discretizing the loss functionals}
\label{sec:dis time}

Finally, we can design implementable algorithms by discretizing the involved (stochastic) integrals and replacing the expectations by Monte Carlo estimators. In that context, we content ourselves with approximating the solution at the grid points defined in \eqref{eq:time grid}, i.e., we seek functions $\widehat{V}_n \in \widehat{\mathcal{H}}$ such that  $\widehat{V}_n(\cdot) \approx V(\cdot, t_n)$, for $n = 0, \ldots, N-1$, and an appropriately chosen function class $\widehat{\mathcal{H}}$ (cf. \Cref{fig: backward iteration scheme}). We note that whilst $\mathcal{H}$ in \eqref{eq: general loss} is understood to contain functions of both space and time, $\widehat{\mathcal{H}}$ comprises functions of space only.  For the forward process \eqref{eq: fordward SDE} we employ the Euler-Maruyama scheme 
\begin{equation}
\label{eq: Euler forward SDE}
    \widehat{X}_{n+1} = \widehat{X}_n + b(\widehat{X}_n, t_n) \Delta t + \sigma(\widehat{X}_n, t_n) \xi_{n+1} \sqrt{\Delta t},
\end{equation}
where $\Delta t = t_{n+1} - t_n$ is the time-step and $\xi_{n+1} \sim \mathcal{N}(0, \operatorname{Id})$ are independent standard normally distributed random variables. It can be shown that $\widehat{X}_n$ approximates $X_{n\Delta t}$ in an appropriate sense as $\Delta t \rightarrow 0$, see \cite{kloeden1992stochastic}. 

For the losses \eqref{eq: residual loss} and \eqref{eq: projection loss} we need to discretize the deterministic and stochastic integrals. Here, we only consider one-step approximations (i.e. we use only one term in the `sum'), but refer to Remark \ref{rem:Jentzen} below for generalizations. As commented on before Remark \ref{rem: L_BSDE to zero}, the integral in the first line of  \eqref{eq: residual loss} can be approximated by either left or right endpoint evaluations of the integrand due to the fact that there is no dependence on $\widehat{V}$. Ultimately, this flexibility allows for the development of both explicit and implicit schemes based on $\mathcal{L}_{\mathrm{BSDE}}$. For the loss $\mathcal{L}_{\mathrm{proj}}$ in \eqref{eq: projection loss}, on the other hand, we must rely on the right endpoint, making use of the availability of $V(X_{t_{n+1}}, t_{n+1}) \approx \varphi(X_{t_{n+1}}, t_{n+1})$ from the previous time-step. 

To simplify notation, we introduce a shorthand notation as follows:
\begin{equation}
\label{eq: shorthand notation h_n}
    {h}_n = h(\widehat{X}_n, t_n, \widehat{V}_n(\widehat{X}_n), \sigma^\top(\widehat{X}_n, t_n)\nabla\widehat{V}_n(\widehat{X}_n)). 
\end{equation}

Starting with $\mathcal{L}_\text{BSDE}$, we obtain the losses 
\begin{equation}
\label{eq: discrete L_BSDE_exp}
\widehat{\mathcal{L}}_\text{BSDE}^\text{exp}(\widehat{V}_n) = \E\left[ \left(\widehat{V}_n(\widehat{X}_n) - {h}_{n+1} \Delta t - \widehat{V}_{n+1}(\widehat{X}_{n+1}) + \sigma^\top(\widehat{X}_n, t_n) \nabla \widehat{V}_n(\widehat{X}_n) \cdot \xi_{n+1}\sqrt{\Delta t}\right)^2  \right],
\end{equation}
and
\begin{equation}
\label{eq: discrete L_BSDE_imp}
    \widehat{\mathcal{L}}_\text{BSDE}^\text{imp}(\widehat{V}_n) = \E\left[ \left(\widehat{V}_n(\widehat{X}_n) - {h}_n \Delta t - \widehat{V}_{n+1}(\widehat{X}_{n+1}) + \sigma^\top(\widehat{X}_n, t_n) \nabla \widehat{V}_n(\widehat{X}_n) \cdot \xi_{n+1}\sqrt{\Delta t}\right)^2  \right].
\end{equation}

Note that $\widehat{\mathcal{L}}_{\text{BSDE}}^\text{exp}$ is explicit in the sense that $h_{n+1}$ only depends on quantities that have already been computed in the previous iteration steps, whereas $\widehat{\mathcal{L}}_{\text{BSDE}}^\text{imp}$ is implicit in the sense that the nonlinear function $h_{n}$ contains the approximating function $\widehat{V}_n$, with respect to which the loss shall be minimized. While implicit numerical schemes may in principle promise improved numerical stability (see e.g. \cite{chassagneux2015numerical}), explicit schemes tend to lead to shorter runtimes. We will demonstrate those computational tradeoffs in multiple numerical examples in \Cref{sec: numerical examples}.

For $\mathcal{L}_\text{proj}$, only the explicit version is available and we obtain
\begin{equation}
\label{eq: discrete L_proj_exp}
\widehat{\mathcal{L}}_{\text{proj}}^\text{exp}(\widehat{V}_n) = \E\left[ \left(\widehat{V}_n(\widehat{X}_n) - h_{n+1} \Delta t - \widehat{V}_{n+1}(\widehat{X}_{n+1})\right)^2\right],
\end{equation}
where the approximation $V(X_{t_{n+1}}, t_{n+1}) \approx \varphi(X_{t_{n+1}}, t_{n+1})$ is used at the right end point of the integral in \eqref{eq: projection loss}.

\begin{remark}[Gradient forcing] 
\label{rem:gradient forcing}
The gradient $\nabla \widehat{V}_n$
is not explicitly part of the objective in \eqref{eq: discrete L_proj_exp}, as $\widehat{V}_{n+1}$ is assumed to be known from the previous iteration. In contrast, both \eqref{eq: discrete L_BSDE_exp} and \eqref{eq: discrete L_BSDE_imp} contain $\nabla \widehat{V}_n$ (note that the short-hand $h_n$ depends on $\nabla \widehat{V}_n$ as well). These differences are partly rooted in the formulas \eqref{eq: residual loss} and \eqref{eq: projection loss}, given that the former is stated in terms of $\nabla \varphi$ rather than $\nabla V$. In challenging (typically high-dimensional) problems where the solution $V$ can only be approximated up to a certain relatively low accuracy, it is plausible that \eqref{eq: discrete L_BSDE_exp}, \eqref{eq: discrete L_BSDE_imp} and \eqref{eq: discrete L_proj_exp} encourage differing trade-offs between the accuracy of $\widehat{V}$ and $\nabla \widehat{V}$; a stark example of this phenomenon will be presented in Section \ref{sec:HJB doublewell}.  
\end{remark}

\begin{remark}[Previous work] 
The loss $\widehat{\mathcal{L}}_{\mathrm{proj}}^\mathrm{exp}$ defined in \eqref{eq: discrete L_proj_exp} has been considered and analyzed extensively in earlier works on the numerical treatment of BSDEs, see  \cite{chessari2021numerical} as well as references therein. A convenient property of this loss is that its explicit nature typically leads to linear regression-type problems that allow for efficient solvers, see \Cref{sec: BSDEs via TTs}. The loss $\widehat{\mathcal{L}}_\mathrm{BSDE}^\mathrm{imp}$ defined in \eqref{eq: discrete L_BSDE_imp} has been considered in \cite{hure2020deep} and \cite{germain2022approximation} in the context of deep learning.
Due to its implicit form, the corresponding algorithms rely on iterative solvers, possibly rendering optimization inefficient.  
The loss $\widehat{\mathcal{L}}_\mathrm{BSDE}^\mathrm{exp}$ defined in \eqref{eq: discrete L_BSDE_exp} is new to the best of our knowledge. It relies on an explicit numerical discretization and, in comparison with $\widehat{\mathcal{L}}_{\mathrm{proj}}^\mathrm{exp}$, enjoys favorable robustness properties when $\varphi \approx V$, see \Cref{sec: robustness properties}.

\end{remark}

\begin{remark}[Discretization of the stochastic integral]
Whilst the Riemann integral in $\mathcal{L}_\mathrm{BSDE}$ can be straightforwardly discretized in multiple different ways,
alternative discretizations of the (It\^{o}) stochastic integral incur (Stratonovich-type) correction terms in the limit as $\Delta t \rightarrow 0$. For example, we may rewrite \eqref{eq: residual loss} in terms of backward integrals (denoted by $\mathrm{d}\overleftarrow{W}$, see \citet[Section 2.7]{kunita2019stochastic}),
\begin{align}
\label{eq:BSDE back}
    \begin{split}
    {\mathcal{L}}_\mathrm{BSDE}(\varphi) &= \E\Bigg[\Bigg(\varphi(X_{t_n}, t_n) - V(X_{t_{n+1}}, t_{n+1}) - \int_{t_n}^{t_{n+1}} h(X_s, s, \varphi(X_s, s), (\sigma^\top \nabla \varphi)(X_s, s))\,\mathrm ds \\
    &\qquad\quad + \int_{t_n}^{t_{n+1}} (\sigma^\top \nabla \varphi)(X_s, s)\cdot \mathrm d \overleftarrow{W}_s - \int_{t_n}^{t_{n+1}} \mathrm{Tr} \left( \sigma \sigma^\top \mathrm{Hess} \, \varphi  \right)(X_s) \, \mathrm{d}s \Bigg)^2 \Bigg],
    \end{split}
\end{align}
where the last term is a correction that has been computed according to equation (2.48) in \citet{kunita2019stochastic}. Discretizing \eqref{eq:BSDE back} then leads to
\begin{align}
    \begin{split}
\label{eq:discrete backwards Ito}
    \mathcal{L}_\mathrm{BSDE,back}^\mathrm{exp}(\widehat{V}_n) = \E\Big[ (\widehat{V}_n(\widehat{X}_n) - \widehat{h}_{n+1} \Delta t - \widehat{V}_{n+1}(\widehat{X}_{n+1}) + \sigma^\top \nabla \widehat{V}_{n+1}(\widehat{X}_{n+1}) \cdot \xi_{n+1}\sqrt{\Delta t} 
    \\
    - \mathrm{Tr} \left( \sigma \sigma^\top \mathrm{Hess} \, \widehat{V}_{n+1}  \right)(\widehat{X}_{n+1}) \Delta t)^2
    \Big],
    \end{split}
\end{align}
where now both the Riemann and the stochastic integral are approximated by the right endpoints, yielding a fully explicit numerical scheme. In our numerical experiments, however, we have not observed clear advantages in terms of accuracy or robustness, and reserve a more careful evaluation for future work.
\end{remark}

\begin{remark}[Losses on the entire time interval]
\label{rem:Jentzen}
Dividing $[0, T]$ into multiple smaller time intervals is attractive since the ensuing subproblems might be easier to solve (cf. \Cref{fig: backward iteration scheme}). However, the continuous-time loss  $\mathcal{L}_\mathrm{BSDE}$ defined in \eqref{eq: residual loss} (and for linear PDEs also the loss $\mathcal{L}_\mathrm{proj}$, see \citet{richter2022robust}) can also be used to approximate the solution on the whole time interval at once, choosing $N = 1$ with $t_0 = 0$ and $t_1 = T$. In order to ensure that the magnitude of the discretization errors remains controlled, one would then need to discretize the integrals on multiple grid points. This, however, automatically leads to implicit schemes and only iterative solvers (or `shooting methods') seem appropriate, see e.g. \cite{weinan2017deep} and \cite{nusken2021interpolating}.
\end{remark}

\begin{remark}[Error propagation]
\label{rem: error propagation}
When dividing the problem into subproblems, on the other hand, we might need to accept a potential propagation of approximation errors, originating from the fact that in implementations the approximation $\widehat{V}_{n+1}(\cdot) \approx V(\cdot, t_{n+1})$ (needed as a `terminal condition' for deriving the individual losses) will usually not be exact, cf. \citet[Section 8.3.3]{gobet2016monte}).
\end{remark}

\section{Robustness properties of the loss functionals}
\label{sec: robustness properties}

In the previous section we have derived numerical schemes for solving parabolic PDEs based on the iterative minimization of appropriate loss functionals. The two losses \eqref{eq: residual loss} and \eqref{eq: projection loss} are both valid in the sense of \eqref{eq: abstract loss property}, i.e. they both yield the solution to the PDE \eqref{eq: definition general PDE} assuming that the expectations involved can be computed exactly and the associated minimization procedure leads to a global minimum. This assumption is not realistic in practice, however. Consequently, this section shall study robustness properties of the losses when the expectations are approximated by a finite sample Monte Carlo estimator. For $\mathcal{L}_\mathrm{BSDE}$ and $\mathcal{L}_{\mathrm{proj}}$ as defined in \eqref{eq: residual loss} and \eqref{eq: projection loss}, the natural estimators are given by
\begin{align}
\label{eq:est proj}
    \begin{split}
    \mathcal{L}_\mathrm{proj}^{(K)}(\varphi) &= \frac{1}{K}\sum_{k=1}^K \Bigg(\varphi(X_{t_n}^{(k)}, t_n) - V(X_{t_{n+1}}^{(k)}, t_{n+1})  \\
    &\qquad\qquad\qquad\qquad\qquad- \int_{t_n}^{t_{n+1}} h(X_s^{(k)}, s, V(X_s^{(k)}, s), (\sigma^\top \nabla V)(X_s^{(k)}, s))\,\mathrm ds \Bigg)^2,
    \end{split}
\end{align}
and
\begin{align}
\label{eq:est BSDE}
    \begin{split}
    \mathcal{L}_\mathrm{BSDE}^{(K)}(\varphi) &= \frac{1}{K}\sum_{k=1}^K \Bigg(\varphi(X_{t_n}^{(k)}, t_n) - V(X_{t_{n+1}}^{(k)}, t_{n+1}) + \int_{t_n}^{t_{n+1}} (\sigma^\top \nabla \varphi)(X_s^{(k)}, s)\cdot \mathrm dW_s^{(k)} \\
    &\qquad\qquad\qquad\qquad\qquad- \int_{t_n}^{t_{n+1}} h(X_s^{(k)}, s, \varphi(X_s^{(k)}, s), (\sigma^\top \nabla \varphi)(X_s^{(k)}, s))\,\mathrm ds \Bigg)^2,
    \end{split}
\end{align}
where $K \in \mathbb{N}$ is the sample size and $(X^{(k)}_s)_{s \in [0,T], k = 1, \ldots, K}$ denote independent and identically distributed copies of the diffusion process \eqref{eq: fordward SDE}, driven by the independent Brownian motions $(W^{(k)}_s)_{s \in [0,T]}$.\par\bigskip

We now aim to study fluctuations of these Monte Carlo approximations at the solution $\varphi = V$ by way of computing the variance associated to \eqref{eq:est proj} and \eqref{eq:est BSDE}. Our results will shed light on the situation when $\varphi \approx V$, which is the relevant regime towards the end of iterative optimization algorithms (such as those considered in Section \ref{sec: BSDEs via TTs}). The following result is in direct correspondence to the observations made in the introduction concerning the differences between the regression objectives \eqref{eq:regression objective} and \eqref{eq:robust loss}.

\begin{proposition}[Expectation and variance of the losses at the solution]
\label{prop: variance of the losses}
Consider the Monte Carlo estimators  \eqref{eq:est proj} and \eqref{eq:est BSDE}, evaluated at the solution $\varphi = V$.
The former has nonzero expectation and variance, namely
\begin{subequations}
\begin{align}
\label{eq: expectation of L_proj}
\E\left[\mathcal{L}_\mathrm{proj}^{(K)}(V)\right] &= \Var\left(\int_{t_n}^{t_{n+1}}(\sigma^\top\nabla V)(X_s, s)\cdot \mathrm dW_s \right),
  \\
  \label{eq: variance of L_proj}
\Var\left(\mathcal{L}_\mathrm{proj}^{(K)}(V)\right) &= \frac{1}{K} \Var\left(\left(\int_{t_n}^{t_{n+1}}(\sigma^\top\nabla V)(X_s, s)\cdot \mathrm dW_s\right)^2 \right).
\end{align}
\end{subequations}
In contrast, the latter is zero almost surely,
\begin{equation}
\label{eq: L_BSDE a.s. zero}
\mathcal{L}_\mathrm{BSDE}^{(K)}(V) = 0, \qquad \mathrm{a.s.},
\end{equation}
implying in particular that both its expectation and variance vanish. 
\end{proposition}

\begin{proof}
Substituting $\varphi = V$ into $\mathcal{L}_{\mathrm{proj}}$ and using the relation \eqref{eq: Ito t_n t_n+1} we see that
\begin{equation}
  \mathcal{L}_\mathrm{proj}(V) = \E\left[\left(\int_{t_n}^{t_{n+1}}(\sigma^\top \nabla V)(X_s, s)\cdot \mathrm dW_s \right)^2 \right] = \Var\left(\int_{t_n}^{t_{n+1}}(\sigma^\top \nabla V)(X_s, s)\cdot \mathrm dW_s  \right),
\end{equation} 
from which \eqref{eq: expectation of L_proj} and \eqref{eq: variance of L_proj} follow directly. For $\mathcal{L}_\mathrm{BSDE}(V)$ note that
\begin{align}
\begin{split}
    V(X_{t_n}^{(k)}, t_n) - V(X_{t_{n+1}}^{(k)}, t_{n+1}) - \int_{t_n}^{t_{n+1}} h(X_s^{(k)}, s, V(X_s^{(k)}, s), (\sigma^\top \nabla V)(X_s^{(k)}, s))\,\mathrm ds \\
    \qquad\qquad\qquad\qquad\qquad\qquad\qquad\qquad\qquad\qquad + \int_{t_n}^{t_{n+1}} (\sigma^\top \nabla V)(X_s^{(k)}, s)\cdot \mathrm dW_s^{(k)} = 0
    \end{split}
\end{align}
holds almost surely for every $k \in \{ 1, \dots, K\}$, see also \eqref{eq: Ito t_n t_n+1}, from which \eqref{eq: L_BSDE a.s. zero} follows immediately and which implies that both the expectation and the variance of the estimator versions are zero.
\end{proof}

\Cref{prop: variance of the losses} shows that the Monte Carlo estimator $\mathcal{L}^{(K)}_\mathrm{proj}$ exhibits noise even at the solution $\varphi = V$. This is an undesirable property, since stochastic optimization algorithms usually degrade in performance in the face of high variance estimators, necessitating smaller learning rates and longer convergence times (\cite{bottou2018optimization}). Close to an optimum,
the presence of noise is likely to induce instabilities and prevent the algorithm from settling down on an accurate solution. 

The estimator $\mathcal{L}^{(K)}_\mathrm{BSDE}$ on the other hand possesses variance zero at the optimum, promising statistical advantages in the optimization routines. In particular, once the approximation is close to the solution the optimization algorithm is expected to be robust and stay close to the optimum. We refer to \Cref{sec: numerical examples} for numerical evidence that illustrates those findings.

To extend our analysis to stochastic gradient descent and its variants, we next study the fluctuations of gradient estimators of the losses. For this, we recall the notion of functional derivative (see e.g. Section 5.2 in \cite{siddiqi1986functional}). For the following definition, we require $\mathcal{L}$ to be defined on a set of functions that is also a vector space. For definiteness, we consider $\mathcal{L}: C^1_b(\mathbb{R}^d \times [0,T]) \rightarrow \mathbb{R}$ (that is, implicitly,  $\mathcal{H} \subset C^1_b(\mathbb{R}^d \times [0,T])$), but note that the boundedness assumption could be relaxed considerably.  

\begin{definition}[G\^{a}teaux derivative]
For $\varphi,\psi \in C^1_b(\mathbb{R}^d \times [0,T])$, 
we say that $\mathcal{L}$ is \emph{G\^{a}teaux differentiable} at $\varphi$ in direction $\psi$ if the mapping
\begin{equation}
\varepsilon \mapsto  \mathcal{L}(\varphi + \varepsilon \psi)
\end{equation}
is differentiable at $\varepsilon=0$. 
The G\^{a}teaux derivative of $\mathcal{L}$ at $\varphi$ in direction $\psi$ is then defined as 
\begin{equation}
    \frac{\delta}{\delta \varphi}\mathcal{L}(\varphi; \psi) := \frac{\mathrm d}{\mathrm d \varepsilon}\Big|_{\varepsilon=0} \mathcal{L}(\varphi+ \varepsilon \psi).
\end{equation}
\end{definition}

We can now investigate the variances of the gradients of the estimated losses. In fact, the following proposition shows that the variance of the gradient of the Monte Carlo estimator of the loss $\mathcal{L}_\mathrm{BSDE}$, as defined in \eqref{eq:est BSDE}, vanishes at the solution $\varphi = V$.

\begin{proposition}[Variance of the gradients at the solution]
\label{prop:variance0}
For every direction $\psi\in C^1_b(\mathbb{R}^d \times [0,T])$ it holds
\begin{equation}
\label{eq:Var0BSDE}
\Var\left(\frac{\delta}{\delta \varphi}\Bigg|_{\varphi = V}\mathcal{L}_\mathrm{BSDE}^{(K)}(\varphi; \psi) \right) = 0.
\end{equation}
\end{proposition}
\begin{proof}
The proof of the identities \eqref{eq:Var0BSDE} and \eqref{eq: variance of grad L_proj} below can be found in Appendix \ref{app:proofs}.
\end{proof}
\begin{remark}[Variance of gradients of the projection loss]
\label{rem:vargradproj}
The remarkable  property \eqref{eq:Var0BSDE} does not hold for the Monte Carlo estimator of the projection loss, defined in \eqref{eq:est proj}. In fact, one can show that
\begin{align}
\label{eq: variance of grad L_proj}    \Var\left(\frac{\delta}{\delta \varphi}\Bigg|_{\varphi  = V}\mathcal{L}_\mathrm{proj}^{(K)}(\varphi; \psi) \right) = \frac{4}{K} \mathbb{E} \left[ (\psi(X_{t_n},t_n))^2 \int_{t_n}^{t_{n+1}} (\sigma^\top \nabla V)(X_s, s))^2 \, \mathrm{d}s\right],
\end{align}
see \Cref{app:proofs} for details. Clearly, the right-hand side of \eqref{eq: variance of grad L_proj} vanishes only in very exceptional cases (for instance, if $\sigma^\top \nabla V \equiv 0$ on $[t_n,t_{n+1}]$).
\end{remark}

\section{Tensor trains for solving BSDEs}
\label{sec: BSDEs via TTs}

\begin{figure}[h]
    \centering
    \begin{minipage}{.3\textwidth}
        \centering
        \begin{tikzpicture}
            \begin{scope}[every node/.style={scale=1,draw,  fill=white}]
                
                \node (A1) at (0,0) {$x$}; 
               	\node [ position=0:-1.5 from A1](A2) {$A$};  	
            \end{scope}
            \begin{scope}[every edge/.style={draw=black,thick}]
            	\path [-] (A1) edge node[midway,left] [above] {$m$} (A2);
            	\path [-] (A2) edge node[midway,left] [above] {$n$} (0:-2.5);
            \end{scope}
        \end{tikzpicture}
        \subcaption{A tensor network representing the order $1$ tensor resulting from the matrix-vector multiplication $Ax = A \circ x \in \mathbb R^n$. }\label{TT:fig:contraction_Ax}
    \end{minipage}
    \begin{minipage}{.68\textwidth}
        \centering
        \begin{tikzpicture}
            \begin{scope}[every node/.style={scale=1,draw,  fill=white}]
                \node (A1) at (0,0) {$A$}; 
               	\node [ position=0:4 from A1](A4) {$U$};  	
               	\node [ position=0:5.5 from A1](A5) {$\Sigma$};  	
               	\node [ position=0:7 from A1](A6) {$V$};  	
            \end{scope}
               	\node [ position=0:2 from A1](A3) {$=$};  	
            \begin{scope}[every edge/.style={draw=black,thick}]
            	\path [-] (A1) edge node[midway,left] [above] {$n$} (0:-1);
            	\path [-] (A1) edge node[midway,left] [above] {$m$} (0:1);
            	
            	\path [-] (A4) edge node[midway,left] [above] {$n$} (0:3);
            	\path [-] (A4) edge node[midway,left] [above] {$r$} (A5);
            	\path [-] (A5) edge node[midway,left] [above] {$r$} (A6);
            	\path [-] (A6) edge node[midway,left] [above] {$m$} (0:8.5);
            \end{scope}
        \end{tikzpicture}
    	\subcaption{Graphical representation of the SVD as a tensor network.}
    \end{minipage}
    \caption{Graphical notation of simple tensors and tensor networks.}
    \label{fig:tikz_simple_network}
\end{figure}
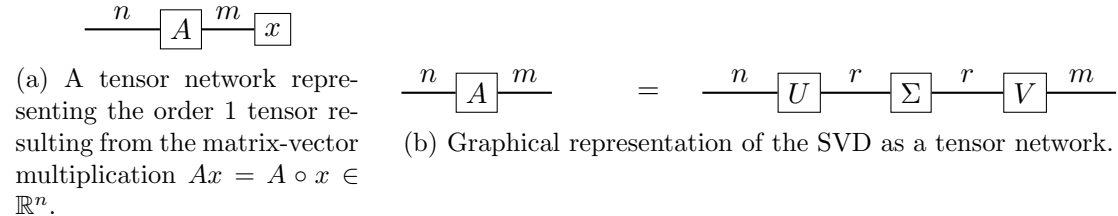

In this section we explain the tensor train approach \citep{oseledets2011tensor} towards representing the  solution $V$ to the PDE \eqref{eq: definition general PDE}, following \citet{holtz2012alternating} and \citet[Chapter 4]{sallandt2022computing}.
We start with the classical model for functions of $d$ real variables, $V: \mathbb{R}^d \rightarrow \mathbb{R}$, using tensorization of functions of one real variable each, $\phi_i:\mathbb{R} \rightarrow \mathbb{R}$.
Given an appropriate set of such functions $\Phi = \{\phi_1, \dots, \phi_m: \mathbb{R} \rightarrow \mathbb{R}\}$,  functions of $d$ real variables can be constructed as\footnote{As hinted at by the choice of variable names, we think of \eqref{eq:V} as an approximation or representation of the solution to the PDE \eqref{eq: definition general PDE}. This correspondence is only vague, as $V$ in \eqref{eq:V} does not depend on time. To be more precise, we should think of \eqref{eq:V} as one instance of $\widehat{V}_i$ within the iterative scheme depicted in Figure \ref{fig: backward iteration scheme}, that is, $t$ is fixed.}
\begin{equation}
\label{eq:V}
    V(x_1, \dots, x_d) = \sum_{i_1 = 1}^m \dots \sum_{i_d = 1}^m c_{i_1, \dots, i_d} \phi_{i_1}(x_1) \dots \phi_{i_d}(x_d).
\end{equation}
The coefficient tensor $c \in \mathbb R^{m \times \dots \times m} = \mathbb R^{m^d}$ is said to be of order $d$ and suffers from the curse of dimensionality: the number of its coefficients grows exponentially,  severely limiting the usage of this classical model in higher dimensions.
To simplify the presentation we introduce the Python-like notation
$c_{i_1, \dots, i_d} = c[i_1, \dots, i_d]$.
In order to introduce the tensor train model we first define the contraction $\circ$ between the last index of a tensor $w_1 \in \mathbb R^{r_1 \times m \times r_2}$ with the first index of another tensor $w_2 \in \mathbb R^{r_2 \times m \times r_3}$ as
\begin{equation}
\label{eq:contraction}
	w = w_1 \circ w_2 \in \mathbb R^{r_1 \times m \times m \times r_3}, \quad w[j_1, i_1, i_2, j_3] = \sum_{j_2 = 1}^{r_2} w_1[j_1, i_1, j_2] w_2[j_2, i_2, j_3].  
\end{equation}
Using this contraction operation, the standard matrix-vector product of $A \in \mathbb{R}^{n \times m}$ and $x \in \mathbb{R}^m$ is given as $A \circ x \in \mathbb{R}^n$. Similarly, the singular value deccomposition (SVD) of $A$ can be written as $A = U \circ \Sigma \circ V$.
Tensors and their contractions can be efficiently visualized as  graphs in which nodes stand for the individual tensors and contractions are denoted as edges between them, cf.\ Figure \ref{fig:tikz_simple_network}. 
We refer to such collections of tensors and contractions between the tensors as \emph{tensor networks}.
Using the $\circ$-notation we can define the tensor~train~representation:
\begin{definition}[Tensor train]\label{def:tensor_train}
Let $c \in \mathbb R^{m \times \dots \times m} = \mathbb{R}^{m^d}$.
A factorization
\begin{equation}
\label{eq:TT rep}
c = u_1 \circ u_2 \circ \dots \circ u_d,
\end{equation}
where $u_1 \in \mathbb R^{m \times r_1}$, $u_i \in \mathbb R^{r_{i-1} \times m \times r_i}$, $2 \leq i \leq d-1$, $u_d \in \mathbb R^{r_{d-1} \times m}$, is called \emph{tensor train representation (TT-representation)} of $c$. 
We refer to the individual tensors $u_i$ as \emph{component tensors}. The tuple $(r_1, \dots, r_{d-1})\in \mathbb{N}^{d-1}$ is referred to as the \emph{representation rank} and is a property of the specific representation \eqref{eq:TT rep}.
    In contrast, the \emph{tensor train rank (TT-rank)} of $c$ is defined as the minimal rank tuple $r = (r_1, \dots, r_{d-1})$ such that there exists a TT-representation of $c$ with representation rank $r$. Here,  minimality of the rank is defined in terms of the partial order relation on $\mathbb N^{d-1}$ given by
\[ s \preceq t \iff s_i \leq t_i \qquad \text{ for all } \,
i \in \{1, \ldots, d-1\},\]
for $s = (s_1, \dots, s_{d-1}), \, t = (t_1, \dots, t_{d-1}) \in \mathbb N^{d-1}$.
\end{definition}
The unique minimal rank tuple can  for example be computed using a multi-linear SVD, cf.\ \cite{holtz2012manifolds}.
Tensor trains have the appealing property that, assuming that the ranks stay bounded, the number of coefficients grows linearly with the spatial dimension.
More precisely, the number of coefficients is at most of order $\mathcal O(d m r)$, where $r = \max_i r_i$. 
The TT-representation is said to be orthogonalized if for some index $\mu$, the component tensors $u_1, \dots, u_{\mu-1}$ are left-orthogonal and $u_{\mu+1}, \dots, u_d$ are right-orthogonal, see e.g.\ \cite{wolf_phd} for a precise definition.
The index $\mu$ associated to the possibly non-orthogonal component tensor is called the core position.
Such an orthogonal representation can be obtained by a sequence of QR-decompositions, and the position of the core can be ``moved'' efficiently (for example, transforming an orthogonal representation with core index $\mu$ into one with core index $\mu + 1$ or $\mu - 1$). We again refer to \cite{wolf_phd} for details.

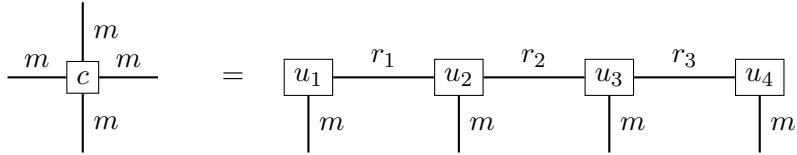
\begin{figure}[h]
    \centering
    \begin{tikzpicture}
        \begin{scope}[every node/.style={scale=1,draw,  fill=white}]
        \node (A1) at (0,0) {$u_1$}; 
        \node (A2) at (2,0) {$u_2$}; 
        \node (A3) at (4,0) {$u_3$}; 
        \node (A4) at (6,0) {$u_4$}; 
        
        \node (A0) at (-3,0) {$c$}; 
        \end{scope}
        \node (B2) at (-1,0) {$=$}; 
        \begin{scope}[every edge/.style={draw=black,thick}]
        	\path [-] (A1) edge node[midway,left,sloped] [above] {$r_1$} (A2);
        	\path [-] (A2) edge node[midway,left,sloped] [above] {$r_2$} (A3);
        	\path [-] (A3) edge node[midway,left,sloped] [above] {$r_3$} (A4);
        	
        	\path [-] (A1) edge node[midway,left] [right] {$m$} +(-90:1);
        	\path [-] (A2) edge node[midway,left] [right] {$m$} +(-90:1);
        	\path [-] (A3) edge node[midway,left] [right] {$m$} +(-90:1);
        	\path [-] (A4) edge node[midway,left] [right] {$m$} +(-90:1);
        	
        	\path [-] (A0) edge node[midway,left] [right] {$m$} +(-90:1);
        	\path [-] (A0) edge node[midway,left] [above] {$m$} +(180:1);
        	\path [-] (A0) edge node[midway,left] [right] {$m$} +(90:1);
        	\path [-] (A0) edge node[midway,left] [above] {$m$} +(0:1);
        \end{scope}
    \end{tikzpicture}
    \caption{An order-$4$ tensor with a possible tensor-train representation.}
    \label{fig:tt_represenation}
\end{figure}

Using the graphical notation for tensor networks we can represent a TT as in Figure \ref{fig:tt_represenation}.

\begin{figure}[h]
    \centering
    \begin{tikzpicture}
        \begin{scope}[every node/.style={scale=1,draw,  fill=white}]
        \node (A1) at (0,0) {$u_1$}; 
        \node (A2) at (2,0) {$u_2$}; 
        \node (A3) at (4,0) {$u_3$}; 
        \node (A4) at (6,0) {$u_4$}; 
        \node (B1) at (0,-1.5) {$\Phi(x_1)$}; 
        \node (B2) at (2,-1.5) {$\Phi(x_2)$}; 
        \node (B3) at (4,-1.5) {$\Phi(x_3)$}; 
        \node (B4) at (6,-1.5) {$\Phi(x_4)$}; 
        
        \end{scope}
        \node (C1) at (-2,-.75) {$V(x)$}; 
        \node (C2) at (-1,-.75) {$=$}; 
        \begin{scope}[every edge/.style={draw=black,thick}]
        	\path [-] (A1) edge node[midway,left,sloped] [above] {$r_1$} (A2);
        	\path [-] (A2) edge node[midway,left,sloped] [above] {$r_2$} (A3);
        	\path [-] (A3) edge node[midway,left,sloped] [above] {$r_3$} (A4);
        	
        	\path [-] (A1) edge node[midway,left] [right] {$m$} (B1);
        	\path [-] (A2) edge node[midway,left] [right] {$m$} (B2);
        	\path [-] (A3) edge node[midway,left] [right] {$m$} (B3);
        	\path [-] (A4) edge node[midway,left] [right] {$m$} (B4);
        	
        \end{scope}
    \end{tikzpicture}
    \caption{A function of $4$ real variables and a TT-representation of its coefficient tensor.}
    \label{fig:functional_tt}
\end{figure}
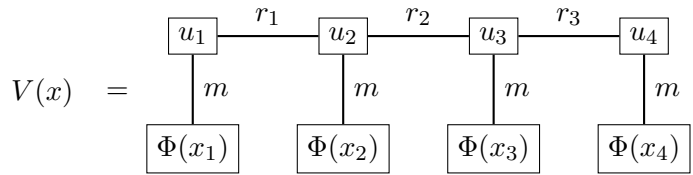

In order to represent functions of several real variables we combine tensor networks with the set of ansatz functions $\Phi = \{ \phi_1, \dots, \phi_m : \mathbb{R} \rightarrow \mathbb{R}\}$, which we shall from now on assume to be linearly independent. 
To simplify the presentation, we overload  the notation and introduce $\Phi : \mathbb R \to \mathbb R^m$, $x \mapsto ( \phi_1(x), \dots, \phi_m(x) )^\top$.
Now we can contract $\Phi(x)$ with the component tensors as shown in Figure \ref{fig:functional_tt} to obtain the function $V$.
We refer to this type of tensor network as a \emph{functional tensor train}.
\subsection{Complexity estimates for common operations}
\label{sec:common ops}
The functional tensor train (Figure \ref{fig:functional_tt}) has appealing properties with respect to the computational cost of common operations such as the evaluation of the function value, its gradient as well as its Laplacian.
All of these operations can be performed in $\mathcal O(d m r^2)$, where $r = \max_i r_i$.
We use the notation from \citet{sallandt2022computing, oster2021approximating}.

{\bfseries Evaluation of $V$:} Let us consider the evaluation of $V$ at a point $x = (x_1, \dots, x_d)^\top \in \mathbb R^d$. 
We first need to compute $\Phi(x_l)$ for $l \in \{1,\ldots,d\}$, which results in $m d$ evaluations of the functions $\phi_i: \mathbb{R} \rightarrow \mathbb{R}$ in total.
We then proceed by performing the contractions in Figure \ref{fig:functional_tt}.
For fixed $l$, the contraction of $\Phi(x_l)$ with the component tensors is carried out from right to left.
Using the $\circ$-notation, this means that
\begin{equation}
\label{eq:V-tt-eval}
    V(x) = u_1 \circ \dots \circ \bigg( \Big(u_{d-1} \circ \big(u_d \circ \Phi(x_d)\big)\Big) \circ \Phi(x_{d-1})\bigg) \circ \dots \circ \Phi(x_1).
\end{equation}
The complexity of the first contraction $v_d = u_d \circ \Phi(x_d)$ is $\mathcal O(rm)$.
The following contraction $u_{d - 1} \circ v_d \circ \Phi(x_{d -1})$ (and also the contractions following thereafter) is of order $\mathcal O(r^2 m)$, while the last contraction is again of order $\mathcal O(rm)$.
This results in a total complexity of order $\mathcal O(dr^2m)$.

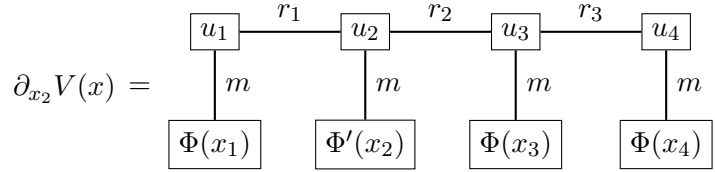
\begin{figure}[h]
    \centering
    \begin{tikzpicture}
        \begin{scope}[every node/.style={scale=1,draw,  fill=white}]
        \node (A1) at (0,0) {$u_1$}; 
        \node (A2) at (2,0) {$u_2$}; 
        \node (A3) at (4,0) {$u_3$}; 
        \node (A4) at (6,0) {$u_4$}; 
        \node (B1) at (0,-1.5) {$\Phi(x_1)$}; 
        \node (B2) at (2,-1.5) {$\Phi'(x_2)$}; 
        \node (B3) at (4,-1.5) {$\Phi(x_3)$}; 
        \node (B4) at (6,-1.5) {$\Phi(x_4)$}; 
        
        \end{scope}
        \node (C1) at (-2,-.75) {$\partial_{x_2} V(x)$}; 
        \node (C2) at (-1,-.75) {$=$}; 
        \begin{scope}[every edge/.style={draw=black,thick}]
        	\path [-] (A1) edge node[midway,left,sloped] [above] {$r_1$} (A2);
        	\path [-] (A2) edge node[midway,left,sloped] [above] {$r_2$} (A3);
        	\path [-] (A3) edge node[midway,left,sloped] [above] {$r_3$} (A4);
        	
        	\path [-] (A1) edge node[midway,left] [right] {$m$} (B1);
        	\path [-] (A2) edge node[midway,left] [right] {$m$} (B2);
        	\path [-] (A3) edge node[midway,left] [right] {$m$} (B3);
        	\path [-] (A4) edge node[midway,left] [right] {$m$} (B4);
        	
        \end{scope}
    \end{tikzpicture}
    \caption{Partial derivative of a $4$-dimensional function in TT-representation.}
    \label{fig:functional_tt_partial}
\end{figure}

{\bfseries Evaluation of $\nabla V$:} In order to compute the gradient $\nabla V$, we start with the evaluation of partial derivatives.
Defining the shorthand $\Phi'(x) = (\phi_1'(x), \dots, \phi_m'(x))^\top$, the partial derivative $\partial_{x_l} V(x_1, \dots, x_d)$ for the case $d = 4$ and $l = 2$ is given in Figure \ref{fig:functional_tt_partial}.
Again, using the $\circ$-notation, but leaving out the brackets for simplicity,\footnote{The order of the contractions is the same as in \eqref{eq:V-tt-eval}, and can be inferred from the dimensions of the individual tensors.} we obtain
\begin{multline}
\label{eq:partial tensornetwork}
    \partial_{x_l} V(x) = u_1 \circ \dots \circ u_{l-1} \circ u_l \circ u_{l+1} \circ \dots \circ u_d \circ
    \\
    \circ \Phi(x_d)\circ \dots \circ \Phi(x_{l+1}) \circ \Phi'(x_l) \circ \Phi(x_{l-1}) \circ \dots \circ \Phi(x_1).
\end{multline}
The complexity of evaluating the partial derivative hence coincides with the complexity of evaluating $V$, i.e.\ $\mathcal O(r^2 m d)$.
In order to compute the gradient, $d$ of these contractions have to be performed, meaning that the complexity of this naive gradient evaluation is $\mathcal O(r^2 m d^2)$.
However, we can improve on that by noticing that in $\partial_{x_l} V(x)$ and $\partial_{x_{l+1}} V(x)$ a number of  contractions appear repeatedly.
Saving these contractions yields an efficient way of computing the gradient.

To make this precise we define the contractions
\begin{equation}
\label{eq:Psiplus}
	\Psi_l^+ (x_{l+1}, \dots, x_d) = u_{l+1} \circ \dots \circ u_d \circ \Phi(x_d) \circ \dots \circ \Phi(x_{l+1}) \in \mathbb R^{r_l},
\end{equation}	
for $l= 1, \ldots, d-1$, as well as
\begin{equation}
\label{eq:Psiminus}
	\Psi_l^- (x_{1}, \dots, x_{l-1}) = u_{1} \circ \dots \circ u_{l-1} \circ \Phi(x_{l-1}) \circ \dots \circ \Phi(x_{1}) \in \mathbb R^{r_{l-1}},
\end{equation}
for $l = 2, \ldots, d$.
In words, $\Psi_l^+$ is the contraction of every component tensor with index larger than $l$, while $\Psi_l^-$ is the contraction of every component tensor with index smaller than $l$.
Using \eqref{eq:Psiplus} and \eqref{eq:Psiminus}, we can rewrite \eqref{eq:partial tensornetwork} in the form
\begin{equation}\label{eq:partial_derivative_psi}
	\partial_{x_l} V(x_1, \dots, x_d) = \big( (\Psi_l^-(x_1, \dots, x_{l-1}) \circ u_l) \circ \Psi_l^+(x_{l+1}, \dots, x_d)\big) \circ \Phi'(x_l),
\end{equation}
for $l = 2, \ldots, d-1$. The edge cases $l=1$ and $l=d$ are covered by
\begin{subequations}
\begin{align}
\partial_{x_1} V(x_1,\ldots,x_d)     = u_l \circ \Psi^+_l.
\end{align}
\end{subequations}
We furthermore observe the recursive identities
\begin{equation}
\label{eq:gradient_psi_recursive}
	\Psi_l^+ = (u_{l+1} \circ \Psi_{l+1}^+) \circ \Phi_{l+1}, \quad \Psi_l^- = \Phi_{l-1} \circ (\Psi_{l-1}^- \circ u_{l-1}),
\end{equation}
omitting the arguments of $\Psi_l^+$ and $\Psi_l^-$.
These allow us to reuse already performed contractions when computing partial derivatives with respect to different variables, and, in combination with  \eqref{eq:partial_derivative_psi}, to arrive at Algorithm \ref{algo:gradient}.
\begin{algorithm}[ht]
\SetAlgoLined
\caption{Computing the gradient of a function $V$ in the TT-format.}\label{algo:gradient}
\SetKwInOut{Input}{Input}
\SetKwInOut{Output}{Output}
\Input{A function $V$ in TT-format (see equation \eqref{eq:V-tt-eval} and Figure \ref{fig:functional_tt}) and its component tensors $u_1, \dots, u_d$, one-dimensional basis functions $\phi_1, \dots, \phi_n$, evaluation point $x = (x_1, \dots, x_d)^\top \in \mathbb R^d$.}
\Output{The gradient $\nabla V(x)$.}
\For{$l = 1, \dots, d-1$}{
    Calculate $\Psi_l^-(x_1, \dots x_{l-1})$ using the recursive formula \eqref{eq:gradient_psi_recursive}.
}
\For{$l = d, \dots, 1$}{
    Calculate $\Psi_l^+(x_{l+1}, \dots, x_d)$ using the recursive formula \eqref{eq:gradient_psi_recursive}. 
}
\For{$l = d, \dots, 1$}{
    Calculate $\nabla V(x) [l] = \frac{\partial V}{\partial x_l}(x)$ using \eqref{eq:partial_derivative_psi}.
}
\end{algorithm}
Note that every micro-step in the algorithm has complexity $\mathcal O(r^2 m)$. Taking into account the for-loops, this amounts to $\mathcal O(d r^2 m)$ and compares favorably to $\mathcal O(d^2 r^2 m)$ for the naive implementation.
Finally, we remark that the Laplacian can be evaluated efficiently in a similar fashion, see \citet{kazeev2012low}.

\subsection{Optimization on the TT manifold}
\label{sec:opt TT manifold}
Now that we have established how to represent the high-dimensional function $V$, we next cover the process of finding optimal coefficients in \eqref{eq:V-tt-eval}, related to the component tensors $u_i$.
We make use of the \emph{alternating least squares} (ALS) algorithm \citep{holtz2012alternating}, which replaces the problem of finding all coefficients at once by a sequence of low-dimensional sub-problems, where only one component tensor is optimized in every iteration.
This is possible due to the multi-linearity of the tensor train representation.

Before moving on to tensor trains (and BSDEs), we first recall the ordinary linear regression problem
\begin{equation}
\label{eq:regression min}
\min_{\varphi \in U} \mathcal{L}^{(K)}(\varphi),
\end{equation}
where $U$ is a finite-dimensional vector space of functions on $\mathbb{R}^d$, and the loss  functional $\mathcal{L}$ takes the form 
\begin{equation}
\label{eq:general regression}
    \mathcal L^{(K)}(\varphi) = \frac{1}{K}\sum_{k = 1}^K | \varphi(x^{(k)}) - y^{(k)} |^2,
\end{equation}
for given samples $x^{(k)} \in \mathbb R^d$ and values (or measurements) $y^{(k)} \in \mathbb R$, $k=1,\ldots,K$.
For a  basis $b_1, \dots, b_N$ of $U$ and a function representation $\varphi(x) = \sum_{i = 1}^N c_i b_i(x)$, the optimal coefficients are given by
\begin{equation}
\label{eq:lin_regression}
    c = (A^\top A)^{-1} A^\top y,
\end{equation}
where $c = (c_1, \dots, c_N)^\top \in \mathbb R^N$, $A = (a_{ij}) \in \mathbb R^{K \times N}$ has coefficients $a_{ij} = b_j(x^{(i)})$, and $y = (y^{(1)}, \dots, y^{(K)})^\top \in \mathbb R^K$, assuming a sufficient amount and nondegeneracy of the samples for $A^\top A$ to be invertible \citep[Chapter 13]{wasserman2004all}.

To explain the extension of ordinary least squares achieved by the ALS algorithm, we first introduce some notation. The set of tensor trains with fixed rank $r \in \mathbb{N}^{d-1}$ is denoted by
\begin{equation}
\label{eq:Mr}
    \mathcal M_r = \{ c = u_1 \circ \dots \circ u_d \ | \ \text{TT-rank}(c)  = r \},
\end{equation}
see Definition \ref{def:tensor_train} for the dimensions of the component tensors $u_i$. It is important to note that $\mathcal{M}_r$ is not a linear subspace of $\mathbb{R}^{m \times \ldots \times m}$ because of the nonlinearity of the contraction operation $\circ$; it is however a submanifold, sometimes referred to as the TT-manifold of rank $r$ \citep{holtz2012manifolds}. We will also need the subsets
\begin{equation}
\label{eq:Mr linear}
    \mathcal{M}_r^l(u_1,\ldots,u_{l-1},u_{l+1},\ldots,u_d) := \left\{c = u_1\circ \ldots \circ u_d: \quad u_l \in \mathbb{R}^{r_{l-1}\times m \times r_l}\right\} \subset \mathcal{M}_r,   
    \end{equation}
obtained from $\mathcal{M}_r$ by fixing the component tensors $u_1,\ldots,u_{l-1},u_{l+1},\ldots,u_d$ and varying only $u_l$. In the following, we will write $\mathcal{M}_r^l$ for notational convenience whenever the particular choices of the fixed component tensors are not relevant for the argument.
A key observation is that $\mathcal{M}_r^l(u_1,\ldots,u_{l-1},u_{l+1},\ldots,u_d)$ is a linear subspace of $\mathbb{R}^{m \times \ldots \times m}$ for any $l \in \{1,\ldots,d\}$ and fixed component tensors, in contrast to $\mathcal{M}_r$. This observation will allow us to efficiently optimize over $\mathcal{M}_r$ using the explicit formula \eqref{eq:lin_regression} by iteratively restricting to the subspaces $\mathcal{M}_r^l$. Associated to the sets of tensor trains defined in \eqref{eq:Mr} and \eqref{eq:Mr linear} and a set of basis functions $\Phi = \{\phi_1,\ldots,\phi_m\}$, we obtain the sets of functional tensor trains $(\mathcal{M}_r,\Phi)$ and  $(\mathcal{M}_r^l(u_1,\ldots,u_{l-1},u_{l+1},\ldots,u_d),\Phi)$ by using the construction principle in \eqref{eq:V-tt-eval} and Figure \ref{fig:functional_tt}. For instance,
\begin{multline}
(\mathcal{M}_r,\Phi) = \Big\{V(x) = u_1 \circ \dots \circ \bigg( \Big(u_{d-1} \circ \big(u_d \circ \Phi(x_d)\big)\Big) \circ \Phi(x_{d-1})\bigg) \circ \dots \circ \Phi(x_1), 
\\
\text{with} \quad u_1 \circ \ldots \circ u_d \in \mathcal{M}_r
\Big\},    
\end{multline}
and similarly for $(\mathcal{M}_r^l,\Phi)$. Clearly, $(\mathcal{M}_r,\Phi)$ and $(\mathcal{M}_r^l,\Phi)$ inherit key properties from $\mathcal{M}_r$ and $\mathcal{M}_r^l$; in particular, $(\mathcal{M}_r^l,\Phi)$ is a linear function space, while $(\mathcal{M}_r,\Phi)$ is merely a (nonlinear) manifold.

We are now interested in replacing the linear space $U$ in \eqref{eq:regression min} by the TT-manifold $(\mathcal{M}_r,\Phi)$, i.e.,
\begin{equation}
\label{eq:regression-tt}
\min_{\varphi \in (\mathcal{M}_r,\Phi)} \mathcal{L}^{(K)}(\varphi),
\end{equation}
for a given rank $r \in \mathbb{N}^{d-1}$ and set of basis functions $\Phi = \{\phi_1,\ldots,\phi_m\}$. The ALS algorithm targets \eqref{eq:regression-tt} by interatively restricting the optimization to $(\mathcal{M}_r^l,\Phi)$, cycling through $l=1,\ldots,d$, while updating the fixed component tensors using the solutions to the sub-regression problems on $(\mathcal{M}^l_r,\Phi)$. More precisely, 
we set $l=1$, start with an arbitrary initialization $u_2,\ldots,u_d$ of the component tensors and consider the  \emph{local regression problem}
\begin{equation}
\label{eq:regression-tt-local1}
\min_{\varphi \in (\mathcal{M}^1_r(u_2,\ldots,u_d),\Phi)} \mathcal{L}^{(K)}(\varphi).
\end{equation}
Crucially, since $(M_r^1(u_2,\ldots,u_d),\Phi)$ is a linear function space, we can apply the solution formula \eqref{eq:lin_regression} to obtain the missing component tensor $\widehat{u}_1$. Proceeding with $l=2$ and the local regression problem (updated using $\widehat{u}_1$)
\begin{equation}
\label{eq:regression-tt-local2}
\min_{\varphi \in (\mathcal{M}^2_r(\widehat{u}_1,u_3,\ldots,u_d),\Phi)} \mathcal{L}^{(K)}(\varphi),
\end{equation}
we perform the update $u_2 \mapsto \widehat{u}_2$. Following this procedure, we update all the component tensors in turn (proceeding again with $\widehat{u}_1$ after $u_d$) until an appropriate termination criterion is met \citep{holtz2012alternating}. While global convergence to the optimum in \eqref{eq:regression-tt} is not guaranteed, it is known that the loss $\mathcal{L}(V)$ decreases at every iteration, see \citet[Section 3.4]{holtz2012alternating}.

We next discuss the details of  an efficient implementation, making use of the contractions $\Psi_l^+$ and $\Psi_l^-$ defined in \eqref{eq:Psiplus} and \eqref{eq:Psiminus}.
Starting with the update of the first component tensor $u_1$ (i.e. $l = 1$ in the discussion above), we observe that $\Psi_1^+$ can be interpreted as an $r_1$-dimensional vector of real-valued functions of $d-1$ variables. Taking the tensor product with the ansatz functions in $\Phi = \{\phi_1,\ldots,\phi_m \}$, we obtain a set of $m \cdot r_1$ functions of $d$ variables.
We formally denote the linear span of functions by $\Phi \otimes \Psi_1^+$, and note that the corresponding coefficients can be stored in a matrix $c \in \mathbb{R}^{m \times r_1}$ whose dimensions equal those of $u_1$. As explained above, solving a regression problem of the type \eqref{eq:lin_regression} then recovers optimal coefficients for the coefficient tensor $u_1$, under the assumption that $u_2, \dots, u_d$ are fixed.

Similarly, local regression problems can be formulated to obtain optimal coefficients for the component tensors $u_l$ for $l \in \{1,\ldots,d\}$, the relevant function spaces being $\Psi_l^- \otimes \Phi \otimes \Psi_l^+$.
As in Section \ref{sec:common ops} (see in particular Algorithm \ref{algo:gradient}), we can use the identities \eqref{eq:gradient_psi_recursive} to compute $\Psi_l^+$ from $\Psi_{l+1}^+$ and vice-versa for $\Psi_l^-$, making the optimization efficient by saving contractions (these saved contractions are commonly referred to as `stacks' within the ALS algorithm).
To summarize, we state the procedure in Algorithm \ref{algo:als_regression}, using the convention that $\Psi_1^- = \Psi_d^+ = \spa \{ 1 \}$ is the space of constant functions.

\begin{algorithm}[ht]
\SetAlgoLined
\caption{Alternating Least Squares (ALS) for regression}\label{algo:als_regression}
\SetKwInOut{Input}{Input}
\SetKwInOut{Output}{Output}
\Input{An order $d$ tensor $u_1 \circ \dots \circ u_d = A \in \mathbb R^{n_1 \times \dots \times n_d}$, sample points $x^{(k)} \in \mathbb R^d$ and data $y^{(k)} \in \mathbb R$.}
\Output{Optimized component tensors $u_1 \circ \dots \circ u_d = A \in \mathbb R^{n_1 \times \dots \times n_d}$.}
\While{not converged}{
\For{$l = d-1$ to $1$}{ \tcp{build stacks}
	Move core position to $l$ (using QR-decompositions).
       
	Compute (and store) $\Psi_l^+(x_{l+1}^{(k)}, \dots, x_d^{(k)})$ for all $k$, using \eqref{eq:gradient_psi_recursive}.
}
    \For{$l = 1$ to $d$}{ \tcp{optimize component tensors}
		Move core position to $l$ (using QR-decompositions).

		Compute (and store) $\Psi_l^-(x_1^{(k)}, \dots, x_{l-1}^{(k)})$ for all $k$, using \eqref{eq:gradient_psi_recursive}.
        
        Solve the regression problem \eqref{eq:regression-tt-local1} for the local basis $\Psi_l^- \otimes \Phi \otimes \Psi_l^+$, using \eqref{eq:lin_regression}.
The solution yields the component tensor $u_l \in \mathbb R^{r_{l-1} \times m_l \times r_{l}}$.
    }
}
\end{algorithm}

To further improve performance, we make two essential modifications of the optimization problem \eqref{eq:regression-tt}.
First, we introduce a regularization term in the loss functional, penalizing the Frobenius-norm of the coefficients in the tensor train (multiplied by a small parameter $\delta > 0$). In the case when $\Phi = \{\phi_1,\ldots,\phi_m\}$ consists of $H^2([a,b])$-orthogonal functions, Parseval's identity implies that this regularizing term can be identified with the norm in  the Sobolev space with dominating mixed smoothness $H^2_{\text{mix}}([a, b]^d)$, see \citet{sickel2009tensor}, so that the loss functional takes the form
\begin{equation}
    \mathcal L^{(K)}(\varphi) = \frac{1}{K}\sum_{k = 1}^K | \varphi(x^{(k)}) - y^{(k)} |^2 + \delta \| \varphi \|_{H^2_{\text{mix}}([a,b]^d)}^2.
\end{equation}
This regularization term can straightforwardly be incorporated in the computational procedure; we simply add $\delta \operatorname{Id}$ to the matrix $A^\top A$ in \eqref{eq:lin_regression}.

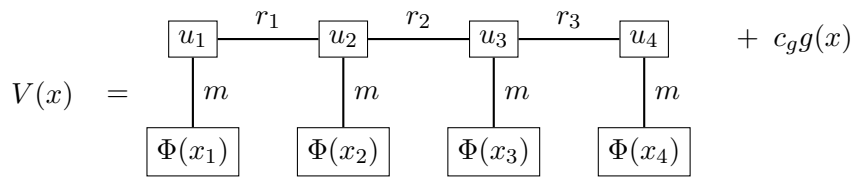
\begin{figure}[h!]
    \centering
    \begin{tikzpicture}
        \begin{scope}[every node/.style={draw,  fill=white}]
        
        \node (A1) at (0,0) {$u_1$}; 
        \node (A2) at (2,0) {$u_2$}; 
        \node (A3) at (4,0) {$u_3$}; 
        \node (A4) at (6,0) {$u_4$}; 
        \node (B1) at (0,-1.5) {$\Phi(x_1)$}; 
        \node (B2) at (2,-1.5) {$\Phi(x_2)$}; 
        \node (B3) at (4,-1.5) {$\Phi(x_3)$}; 
        \node (B4) at (6,-1.5) {$\Phi(x_4)$}; 
        
        \end{scope}
        \node (C1) at (-2,-.75) {$V(x)$}; 
        \node (C2) at (-1,-.75) {$=$}; 
        
        \node (C2) at (8,0) {$+\,\,\, c_g g(x)$}; 
        \begin{scope}[every edge/.style={draw=black,thick}]
        	\path [-] (A1) edge node[midway,left,sloped] [above] {$r_1$} (A2);
        	\path [-] (A2) edge node[midway,left,sloped] [above] {$r_2$} (A3);
        	\path [-] (A3) edge node[midway,left,sloped] [above] {$r_3$} (A4);
        	\path [-] (A1) edge node[midway,left] [right] {$m$} (B1);
        	\path [-] (A2) edge node[midway,left] [right] {$m$} (B2);
        	\path [-] (A3) edge node[midway,left] [right] {$m$} (B3);
        	\path [-] (A4) edge node[midway,left] [right] {$m$} (B4);
        	
        \end{scope}
    \end{tikzpicture}
    \caption{Graphical representation of $V: \mathbb R^4 \to \mathbb R$, with $g$ included in the model space.}
    \label{fig:valuefun3}
\end{figure}

The second modification slightly alters the model space $(\mathcal{M}_r,\Phi)$. 
Oftentimes the terminal condition $g$ cannot be represented exactly (or conveniently) within $(\mathcal{M}_r,\Phi)$ due to the specific tensor train structure. Addressing this, we can simply add $g$ to the model and obtain a representation of $V$ as in Figure \ref{fig:valuefun3} (in the same manner, other functions of particular relevance could be added to the model space as well). 
The modification of the ALS algorithm for this modified model is straightforward, as the local basis $\Psi_l \otimes \Phi \otimes \Psi_l$ can be modified to $(\Psi_l \otimes \Phi \otimes \Psi_l) \oplus g$, thus increasing the dimension of the local regression problem to $r_l m r_{l+1} + 1$.

\begin{remark}
In \citet{gotte2021block} as well as \citet{trunschke2021convergence},
sparsity promoting 
modifications of \eqref{eq:regression-tt} have been considered, the latter of which is based on an $L^1$ rather than an $L^2$ penalty. These variations could reduce the number of samples needed for accurate estimation, further boosting the performance of the algorithm.
\end{remark}

\label{sec: handling implicit regression}

\subsection{Handling gradient-dependent loss functionals}
\label{sec: handling L_BSDE_exp}
The loss functionals $\widehat{\mathcal{L}}_\text{BSDE}^\text{exp}$ and $\widehat{\mathcal{L}}_\text{BSDE}^\text{imp}$ (see equations \eqref{eq: discrete L_BSDE_exp} and \eqref{eq: discrete L_BSDE_imp}) are not of the form \eqref{eq:general regression} because of their dependence on $\nabla V$, and therefore cannot directly be optimized using Algorithm \ref{algo:als_regression}.
However, their inherently similar structure can still be leveraged to efficiently approximate minimizers in a TT-setting. More specifically, we can replace \eqref{eq:general regression} by
\begin{equation}
\label{eq:regression with nabla}
    \mathcal L^{(K)}(\varphi)  = \frac{1}{K}\sum_{k = 1}^K | \varphi(x^{(k)}) + \nabla \varphi(x^{(k)}) \cdot \Xi^{(k)} - y^{(k)} |^2,
\end{equation}
with $x^{(k)} \in \mathbb R^d$, $\Xi^{(k)} \in \mathbb R^d$, and $y^{(k)} \in \mathbb R$ for $1 \leq k \leq K$. In the case of $\widehat{\mathcal{L}}_\text{BSDE}^\text{exp}$, for instance, we see that setting $y^{(k)} = h^{(k)}_{n+1} \Delta t + \widehat{V}_{n+1}(\widehat{X}_{n+1})$ and $\Xi^{(k)} = \sqrt{\Delta t} \sigma(\widehat{X}_n^{(k)},t_n) \xi_{n+1}^{(k)}$ recovers the standard Monte Carlo estimator for \eqref{eq: discrete L_BSDE_exp}. 

As in Section \ref{sec:opt TT manifold}, we first cover the problem of minimizing \eqref{eq:regression with nabla} over a finite-dimensional linear space of (differentiable) functions on $\mathbb{R}^d$, see the formulation in \eqref{eq:regression min}. 
Inserting a representation $\varphi(x) = \sum_{i = 1}^N c_i b_i(x)$ in terms of smooth basis functions, the solution again takes the form \eqref{eq:lin_regression}, $c = (A^\top A)^{-1} A^\top y$, with a modified matrix $A \in \mathbb R^{N \times K}$,
\begin{equation}\label{eq:linear_semi_matrixentries}
A_{ik} = b_i(x^{(k)}) + \sum_{j = 1}^d (\partial_{x_j} b_i(x^{(k)})[j]) \Xi^{(k)}[j].
\end{equation}
As before we use the Python-like notation to access entries of a vector. 
In order to obtain an efficient algorithm saving contractions, we set out to find a recursive representation of function (and gradient) evaluations.
The first term in \eqref{eq:linear_semi_matrixentries} can be handled as in Section \ref{sec:opt TT manifold}, using the operations defined in \eqref{eq:Psiplus} and \eqref{eq:Psiminus}. To handle the second term, we define
\begin{subequations}
\begin{align}
\Theta_l^-(x_1, \dots, x_{l-1}, \xi) &= \sum_{j = 1}^{l-1} \partial_{x_j} \Psi_l^-(x_1, \dots, x_{l-1}) \xi[j],\\
\Theta_l^+(x_{l+1}, \dots, x_{d}, \xi) &= \sum_{j = l+1}^{d} \partial_{x_j} \Psi_l^+(x_{l+1}, \dots, x_{d}) \xi[j]
\end{align}
\end{subequations}
and notice that
\begin{align}
\Theta_l^-(x_1, \dots, x_{l-1}, \xi) = \Theta_{l-1}^-(x_1, \dots, x_{l-2}, \xi) \circ u_{l-1} \circ \Phi(x_{l-1})
\nonumber
\\
+ \Psi(x_1, \dots, x_{l-2}) \circ u_{l-1} \circ \Phi(x_{l-1}) .
\label{eq:gradient_theta_recursive}
\end{align}
Finally, building on \eqref{eq:Psiplus}-\eqref{eq:Psiminus} and \eqref{eq:gradient_theta_recursive}, the identity
\begin{align}
\sum_{j = 1}^d \partial_{x_j} \varphi(x) \xi[j] & = \Theta_l^-(x_1, \dots, x_{l-1}, \xi) \circ \Psi_l^+(x_{l+1}, \dots, x_d) \circ \Phi(x_l) \nonumber
 \\
&+ \Psi_l^-(x_1, \dots, x_{l-1})  \circ  u_l \circ \Psi_l^+(x_{l+1}, \dots, x_d) \circ  \Phi'(x_l) \cdot \xi[l] 
 \nonumber
 \\
&+ \Psi_l^-(x_1, \dots, x_{l-1}) \circ u_l \circ \Theta_l^+(x_{l+1}, \dots, x_d, \xi) \circ \Phi(x_l) \label{eq:sum_of_gradient_fast} 
\end{align}
yields an efficient way to evaluate $\sum_{j=1}^d \partial_{x_j} \varphi(x) \xi[j]$.
We summarize the resulting procedure in Algorithm \ref{algo:als_semi}.
\begin{algorithm}[ht]
\SetAlgoLined
\caption{Alternating Least Squares (ALS) for losses involving gradients}\label{algo:als_semi}
\SetKwInOut{Input}{Input}
\SetKwInOut{Output}{Output}
\Input{An order $d$ tensor $u_1 \circ \dots \circ u_d = A \in \mathbb R^{n_1 \times \dots \times n_d}$, sample points $x^{(k)} \in \mathbb R^d$ and data $y^{(k)} \in \mathbb R$.}
\Output{Optimized component tensors $u_1 \circ \dots \circ u_d = A \in \mathbb R^{n_1 \times \dots \times n_d}$.}
\While{not converged}{
\For{$l = d-1$ to $1$}{ \tcp{build stacks}
	Move core position to $l$ (using QR-decompositions).
       
	Compute (and store) $\Psi_l^+(x_{l+1}^{(k)}, \dots, x_d^{(k)})$ for all $k$, using \eqref{eq:gradient_psi_recursive}.
	
	Compute (and store) $\Theta_l^+(x_{l+1}^{(k)}, \dots, x_d^{(k)}, \xi^{(k)})$ for all $k$, using \eqref{eq:gradient_theta_recursive}.
}
    \For{$l = 1$ to $d$}{ \tcp{optimize component tensors}
		Move core position to $l$ (using QR-decompositions).

		Compute (and store) $\Psi_l^-(x_1^{(k)}, \dots, x_{l-1}^{(k)})$ for all $k$, using \eqref{eq:gradient_psi_recursive}.
		
		Compute (and store) $\Theta_l^-(x_1^{(k)}, \dots, x_{l-1}^{(k)}, \xi^{(k)})$ for all $k$, using \eqref{eq:gradient_theta_recursive}.
        
        Solve the local regression problem for \eqref{eq:regression with nabla} and the local basis $\Psi_l^- \otimes \Phi \otimes \Psi_l^+$, using \eqref{eq:linear_semi_matrixentries}.
The solution yields the component tensor $u_l \in \mathbb R^{r_{l-1} \times m_l \times r_{l}}$.
    }
}
\end{algorithm}

\section{Numerical examples}
\label{sec: numerical examples}

\begin{algorithm}[ht]
\SetAlgoLined
\caption{Approximating the solution to the PDE \eqref{eq: definition general PDE}}\label{algo: PDE approximation}
\SetKwInOut{Input}{Input}
\SetKwInOut{Output}{Output}
	\Input{Initial parametric choice for the functions $\widehat{V}_n$, for $n \in \{0, \dots, N-1 \}$, e.g. tensor~trains with specified ranks and polynomial degrees.}
\Output{Approximation of $V(\cdot, t_n) \approx \widehat{V}_n$ along the trajectories for $n \in \{0, \dots, N-1 \}$.}
Simulate $K$ samples of the discretized SDE \eqref{eq: Euler forward SDE}.  

Choose $\widehat{V}_N = g$.

\For{$n = N - 1$ to $0$}{
	Choose one of the losses \eqref{eq: discrete L_BSDE_exp}, \eqref{eq: discrete L_BSDE_imp} or \eqref{eq: discrete L_proj_exp}.
       
	Minimize this quantity (explicitly or by iterative schemes, see Algorithms \ref{algo:gradient}-\ref{algo:als_semi}).
	
	Set $\widehat{V}_n$ to be the minimizer.
}

\end{algorithm}

In this section, we extend the numerical examples from \citep{richter2021solving} by incorporating the explicit robust loss defined in \eqref{eq: discrete L_BSDE_exp} and comparing it to the other losses from \Cref{sec: BSDE}. As in \citep{richter2021solving}, we  contrast tensor trains with neural networks for function approximation. We refer to \Cref{app: implementation details} for implementation details and the definition of the error metrics (RMSE, relative error and PDE loss). Here, we only highlight the fact that RMSE and relative error quantify the accuracy of the approximation in terms of function values, whereas the PDE loss is also sensitive to discrepancies in the derivatives. The code can be found at \url{https://github.com/lorenzrichter/PDE-backward-solver}. For convenience, we summarize an overview of the general method in Algorithm \ref{algo: PDE approximation}, as already shown in \citep{richter2021solving}.

\subsection{Hamilton-Jacobi-Bellman equation}

Associated to (stochastic) optimal control problems,
the Hamilton-Jacobi-Bellman (HJB) equation is a PDE for the value function, representing the minimal cost-to-go from which the optimal control policy can be deduced. As suggested in \cite{weinan2017deep}, we consider the specific form
\begin{subequations}
\begin{align}
    \left(\partial_t + \Delta \right)V(x, t) - |\nabla V(x, t)|^2 &= 0, \\
    V(x, T) &= g(x),
\end{align}
\end{subequations}
with $g(x) = \log\left(\frac{1}{2} + \frac{1}{2}|x|^2 \right)$, leading to
\begin{equation}
    b = \mathbf{0},\quad \sigma = \sqrt{2} \, \mathrm{Id}_{d \times d},\quad h(x, s, y, z) = -\frac{1}{2}|z|^2,
\end{equation}
using the notation from Section \ref{sec:setting}. A reference solution can be obtained from
\begin{equation}
\label{eq: HJB reference solution}
    V(x, t) = -\log \E \left[e^{-g(x + \sqrt{T- t}\sigma \xi)} \right],
\end{equation}
where $\xi \sim \mathcal{N}(\mathbf{0}, \mathrm{Id}_{d\times d})$ is a normally distributed random variable, see e.g. Appendix D.1 in \cite{richter2021solving} for a derivation of this formula.

\begin{figure}[h!]
\vskip 0.2in
\begin{center}
\centerline{\includegraphics[width=1\columnwidth]{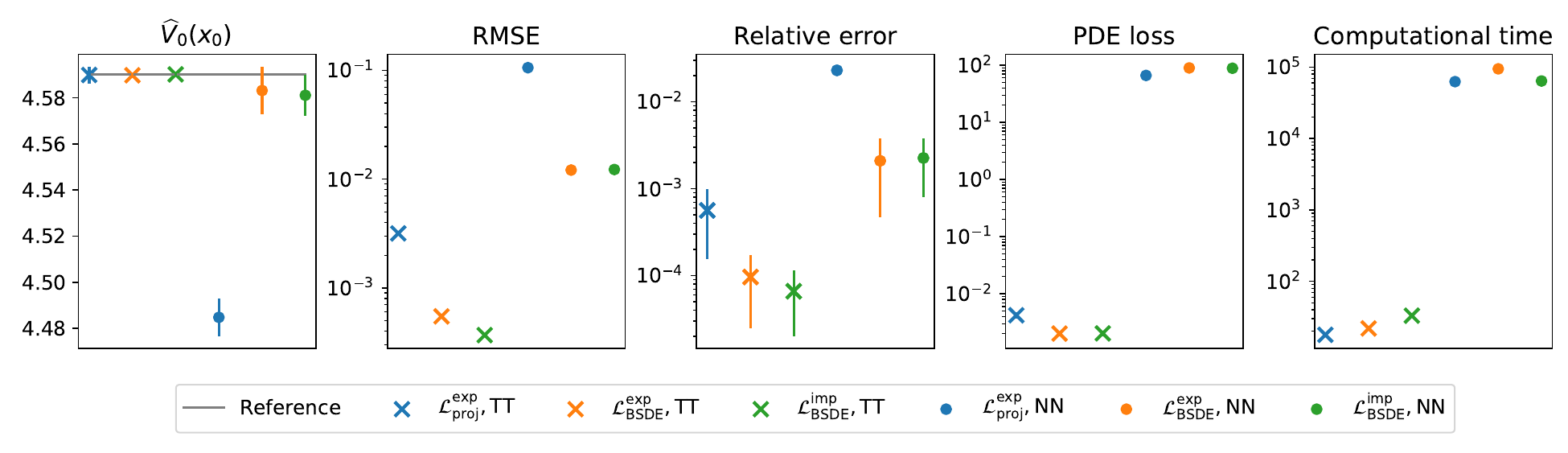}}
\caption{We compare different loss functions for a $100$-dimensional HJB example, either relying on tensor trains or on neural networks. For computational details we refer to \Cref{app: implementation details}.}
\label{fig: HJB d = 100 stats}
\end{center}
\vskip -0.2in
\end{figure}

In our experiments we set $d = 100$, $T = 1$, $\Delta t = 0.01$, $x_0 = (0, \dots, 0)^\top$ and $K = 2000$. Throughout, we complete 100 independent runs of  Algorithm \ref{algo: PDE approximation}, each with different realizations of the Brownian increments and with different randomly initialized tensor trains and neural networks.
For the tensor train model we first compare different polynomial degrees as well as ranks and observe that choosing the polynomial degree to be $0$ and the rank to be $1$ yields the best results, in terms of relative error, RMSE and PDE loss. The neural network settings are summarized in Appendix C in \citep{richter2021solving}. We display the means and standard deviations of the runs in \Cref{fig: HJB d = 100 stats} (if no error bars are visible, this means that the standard deviations of the results are very small in comparison to the resolution). The approximations based on tensor trains  are obtained  significantly faster than those using neural networks, while at the same time being substantially more accurate. Comparing the different losses from \Cref{sec: BSDE} we see that the BSDE losses lead to a notable increase in performance at the cost of a higher computational budget. The explicit and implicit versions of the BSDE loss perform similarly, in this case with small advantages for the implicit version.

To confirm these results, we plot evaluations of the (approximated) value function along two realizations of the forward process in Figures \ref{fig: HJB trajectories, d = 10} and \ref{fig: HJB trajectories}, for dimensions $d=10$ and $d=100$, respectively, comparing the tensor train and neural network approximations, both relying on the implicit BSDE loss. We observe that especially in high dimensions the tensor train approximation agrees more with the reference solution than the neural network approximation. Finally, we plot the mean relative error as a function of time in Figure \ref{fig: HJB mean relative error}, which again shows that the tensor train model outperforms the neural network approach.\par\bigskip

\begin{figure}[h!]
\vskip 0.2in
\begin{center}
\centerline{\includegraphics[width=0.65\columnwidth]{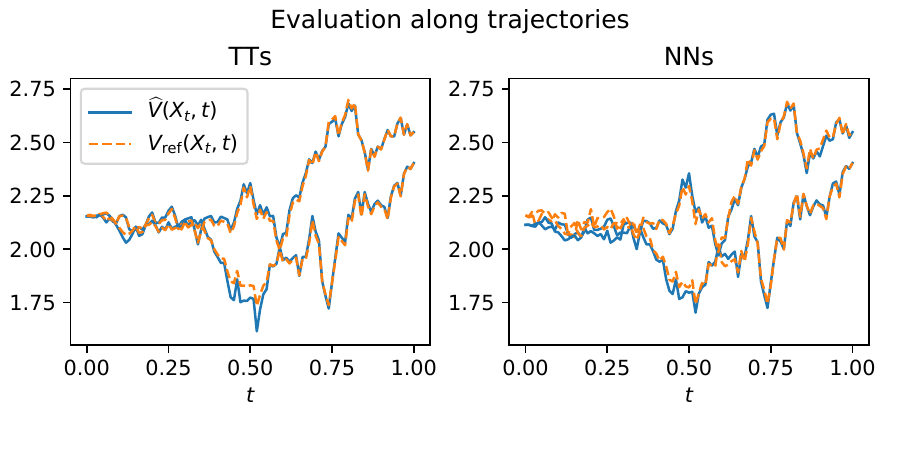}}
\caption{Reference solutions compared with tensor train and neural network approximations using the implicit BSDE loss along two trajectories in $d=10$.}
\label{fig: HJB trajectories, d = 10}
\end{center}
\vskip -0.2in
\end{figure}

\begin{figure}[h!]
\vskip 0.2in
\begin{center}
\centerline{\includegraphics[width=0.75\columnwidth]{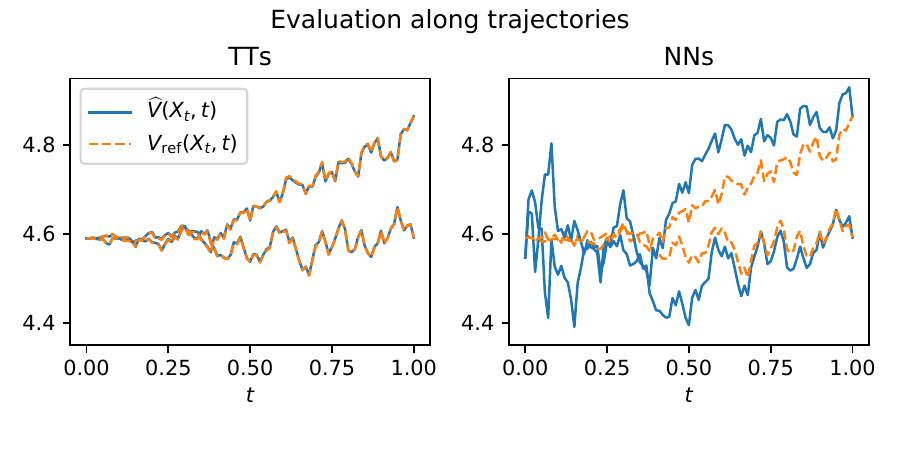}}
\caption{Reference solutions compared with tensor train and neural network approximations using the implicit BSDE loss along two trajectories in $d=100$.}
\label{fig: HJB trajectories}
\end{center}
\vskip -0.2in
\end{figure}

\begin{figure}[h!]
\vskip 0.2in
\begin{center}
\centerline{\includegraphics[width=.55\columnwidth]{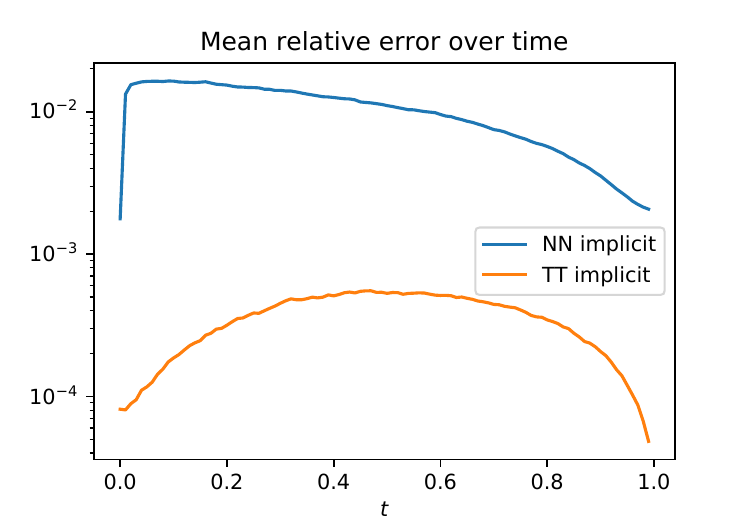}}
\caption{Mean relative error over time for the tensor train and neural network attempts, relying on the implicit BSDE loss.}
\label{fig: HJB mean relative error}
\end{center}
\vskip -0.2in
\end{figure}

\begin{figure}[h!]
\vskip 0.2in
\begin{center}
\centerline{\includegraphics[width=0.6\columnwidth]{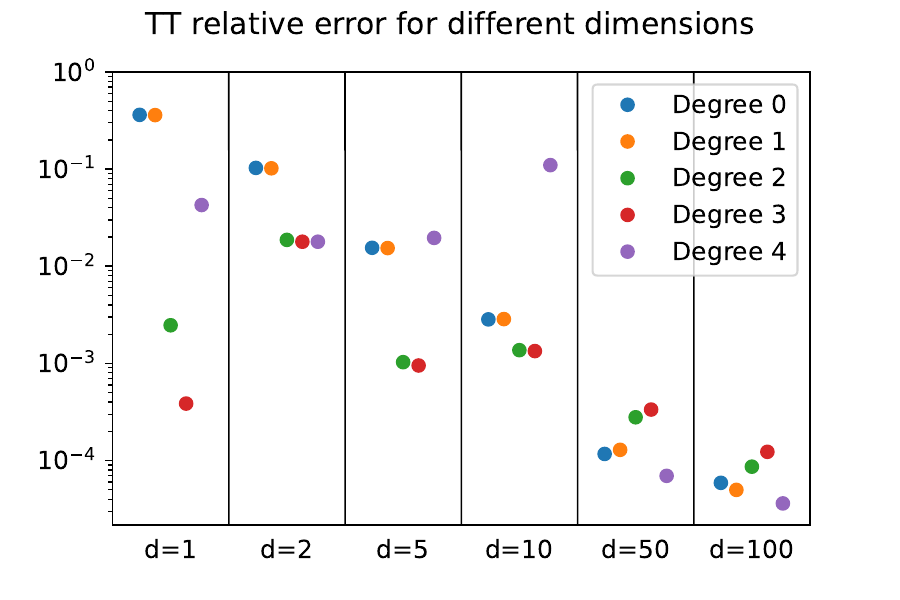}}
\caption{We display relative errors achieved with the implicit BSDE loss and the tensor train approach for varying dimensions and different polynomial degrees. Since the polynomial degree necessary for achieving a good performance seems to decrease with growing state space dimension, we hypothesize a \textit{blessing of dimensionality}.}
\label{fig: blessing of dimensionality}
\end{center}
\vskip -0.2in
\end{figure}

As observed in \citep{richter2021solving}, the model simplicity of the tensor train approach is rather surprising (polynomial degree $0$, i.e. constant ansatz functions), and therefore the dependence of the polynomial degree for different problem dimensions was studied in more detail. In \Cref{fig: blessing of dimensionality} we display relative errors achieved with the tensor train approach using the implicit BSDE loss for varying dimensions and different polynomial degrees. Interestingly, we can see that the problem appears to become easier with growing dimensions and that only for smaller dimensions large polynomial degrees are beneficial. We hypothesize that the observed effect can be attributed to a \emph{blessing of dimensionality}, a phenomenon known from the theory of interacting particle systems (``propagation of chaos'', see \cite{sznitman1991topics}), where in various scenarios, the joint distribution of a large number of particles tends to approximately factorize as the number of particles increases (that is, as the dimensionality of the joint state space grows large). Similar effects were also reported in \cite{bayer2021pricing} (Figure 3) and \cite{khoromskij2012tensors} (Section 1.3), but a theoretical understanding is still lacking. It is plausible that approximate factorizations are relevant for high-dimensional PDEs and that tensor methods are useful (i) to detect those factorizations and (ii) to exploit them, while the black-box nature of the neural networks does not reveal such properties.

\subsection{HJB with double well dynamics}
\label{sec:HJB doublewell}
Let us continue with another HJB example, which was proposed by \citet{nusken2020solving}. In this example the forward SDE is nonlinear, which is likely to result in more complicated structures of the PDE solution. We consider
\begin{subequations}
\begin{align}
    \left(\partial_t + L \right)V(x, t) - \frac{1}{2}|(\sigma^\top \nabla V)(x, t)|^2 &= 0, \\
    V(x, T) &= g(x),
\end{align}
\end{subequations}
with $L$ as in \eqref{eq: infinitesimal generator}, where now the drift is given as the gradient of a double well potential,
\begin{equation}
    b = -\nabla \Psi, \qquad \Psi(x) = \sum_{i,j=1}^d C_{ij}(x_i^2 - 1)(x_j^2 - 1),
\end{equation}
with the matrix $C \in \mathbb R^{d\times d}$ assumed to be positive definite. The terminal condition is given by $g(x) = \sum_{i=1}^d \nu_i(x_i - 1)^2$ with parameters $\nu_i > 0$. 

\begin{figure}[h!]
\vskip 0.2in
\begin{center}
\centerline{\includegraphics[width=0.8\columnwidth]{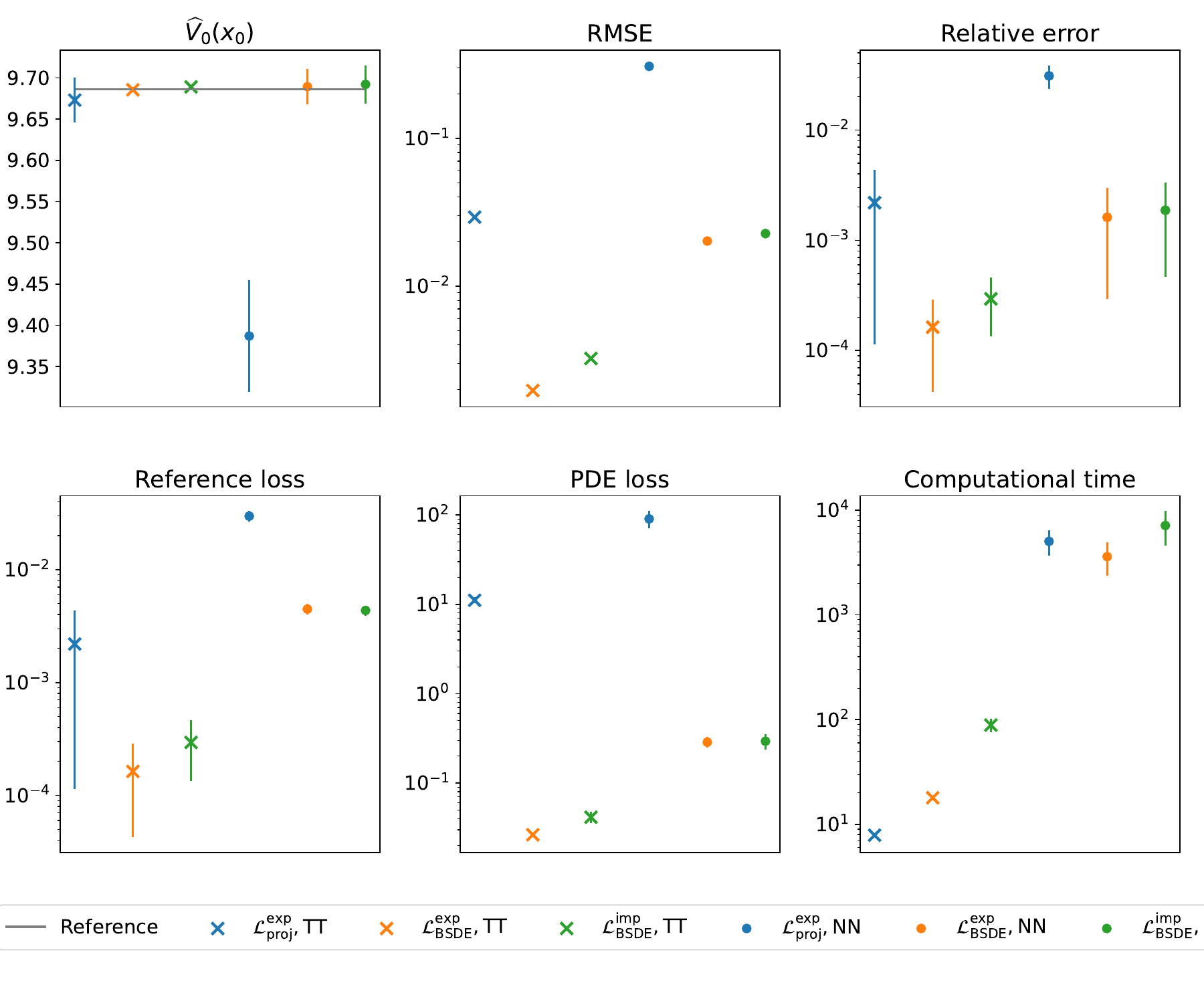}}
\caption{We compare different loss functions for a $50$ dimensional HJB example that relates to a nonlinear SDE with the drift given by the negative gradient of a multidimensional extension of a double well potential, either relying on tensor trains or neural networks. For computational details of the evaluation we refer to \Cref{app: implementation details}.}
\label{fig: double well diagonal metrics 2}
\end{center}
\vskip -0.2in
\end{figure}

We first consider a factorized scenario, with a diagonal matrix $C = 0.1 \,\mathrm{Id}$, allowing us to estimate the optimal TT-rank to be $2$. Further, we set $\sigma = \sqrt 2\mathrm{Id}$, $T = 0.5$, $d = 50$, $\Delta t = 0.01$, $K = 2000$, $\nu_i = 0.05$ and $x_0 = (-1, \dots, -1) \in \mathbb R^d$. For the tensor train approximation we choose the polynomial degree to be $3$. Using the fact that the solution factorizes,  we can compute a low-dimensional reference solution based on  finite differences. We display the results of Algorithm \ref{algo: PDE approximation} in \Cref{fig: double well diagonal metrics 2}, again comparing tensor trains with neural networks as well as the three main losses from \Cref{sec: BSDE}. As before, the tensor trains are both faster and more accurate than neural networks. We further observe that the explicit BSDE loss yields slightly better results than its implicit counterpart, while being almost one order of magnitude faster.

\subsubsection*{Double well with interacting dimensions}

\begin{figure}[h!]
\centering
\begin{subfigure}[b]{1\textwidth}
\includegraphics[width=1\linewidth]{{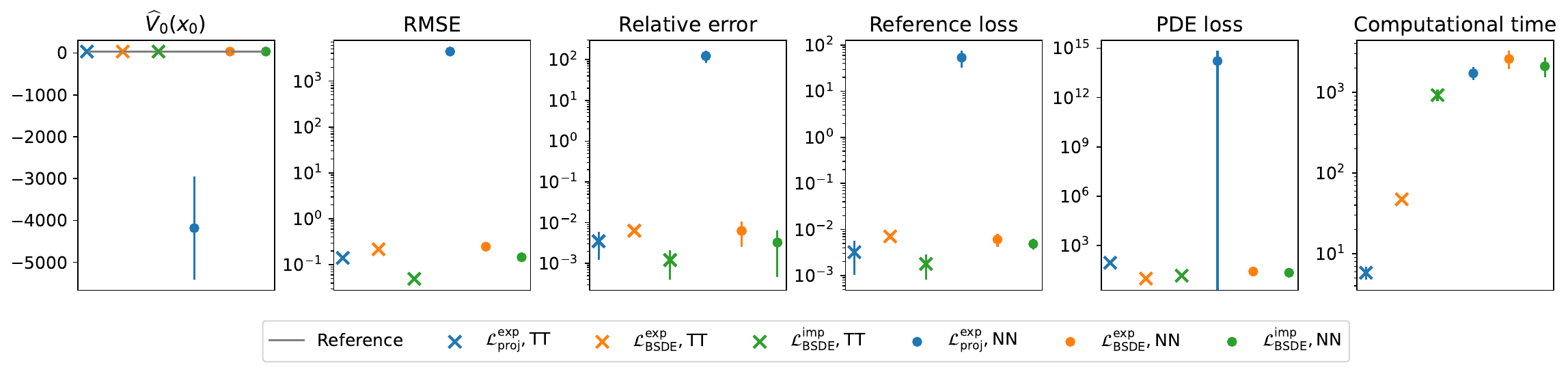}}
   \caption{} 
\end{subfigure}

\begin{subfigure}[b]{1\textwidth}
   \includegraphics[width=1\linewidth]{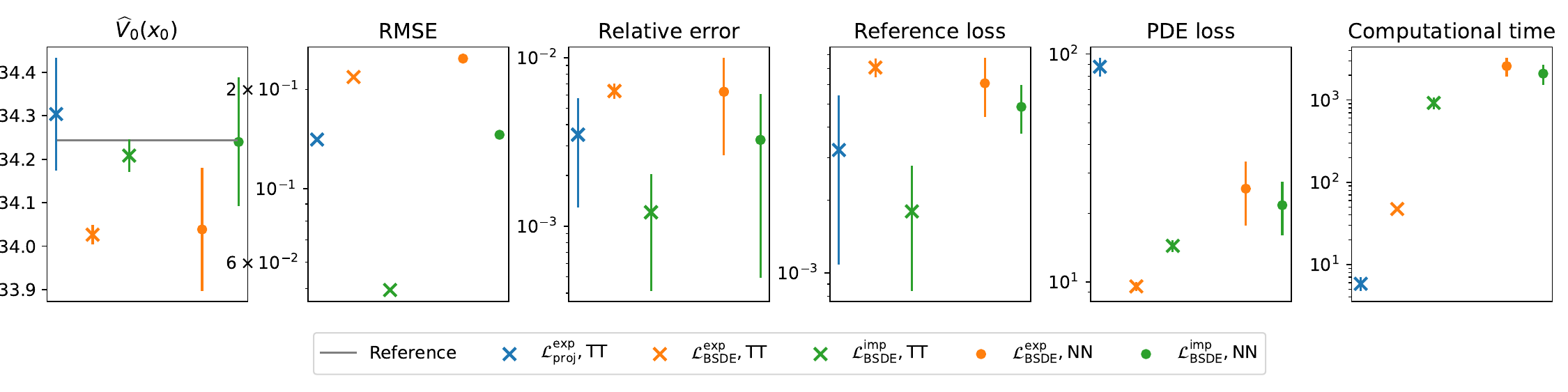}
   \caption{}
\end{subfigure}

\caption[]{We compare different loss functions for a $20$ dimensional HJB example that relates to a nonlinear SDE with the drift given by the negative gradient of a multidimensional extension of a double well potential with interacting dimensions, either relying on tensor trains or neural networks. Subfigure (a) shows that the implicit BSDE loss does not yield satisfactory results in combination with neural networks. To allow for a more detailed comparison between the remaining approaches, these results are discarded in subfigure (b).
For computational details of the evaluation we refer to 
\Cref{app: implementation details}.}
\label{fig: double well non-diagonal}
\end{figure}

We continue by introducing nondiagonal elements in the matrix $C$, expecting that higher TT-ranks will be needed to cope with the coupled nature of the problem. We set $d = 20$, $T = 0.3$, $\nu_i = 0.5$ and $C = \mathrm{Id} + \xi_{ij}$, where $\xi_{ij} \sim \mathcal N(0, 0.01)$ are sampled once at the beginning of the experiment -- all other constants are kept the same as before. We now compute a reference solution by using a Monte Carlo estimate for 
\begin{equation}
\label{eq: HJB double well reference solution}
    V(x, t) = -\log\E\left[e^{-g(X_T)} \Big| X_t = x \right]
\end{equation}
based on $10^7$ samples, obtaining $V(x_0, 0) \approx 34.2687$. For the explicit losses we do not specify the tensor train ranks and instead let the rank-adaptive solver find them. These ranks are mostly between $4$ and $6$. For the implicit losses the ranks are growing within the iterations if we do not cap them. Motivated by the results for the explicit case we cap the ranks at $r_i \leq 6$. We choose polynomial degree $7$ and obtain the results displayed in \Cref{fig: double well non-diagonal}.

In this example, the tensor trains are overall competitive with the neural networks in terms of accuracy, whilst offering shorter computing times. Comparing the results for RMSE and PDE loss reveals an interesting phenomenon: For the regression-based tensor train approaches, losses that explicitly incorporate gradient information ($\mathcal{L}^{\mathrm{exp}}_{\mathrm{BSDE}}$ and $\mathcal{L}^{\mathrm{imp}}_{\mathrm{BSDE}}$) perform better in terms of PDE loss, but worse in terms of RMSE. This observation is in line with Remark \ref{rem:gradient forcing}, and this trade-off might inform the choice of method depending on the application: In optimal control settings, for example, it is often the case $\nabla V$ is directly related to the optimal control, and hence of primary interest (rather than $V$, which might turn out secondary).

\subsection{Cox–Ingersoll–Ross model}
\begin{figure}[h!]
\vskip 0.2in
\begin{center}
\centerline{\includegraphics[width=1\columnwidth]{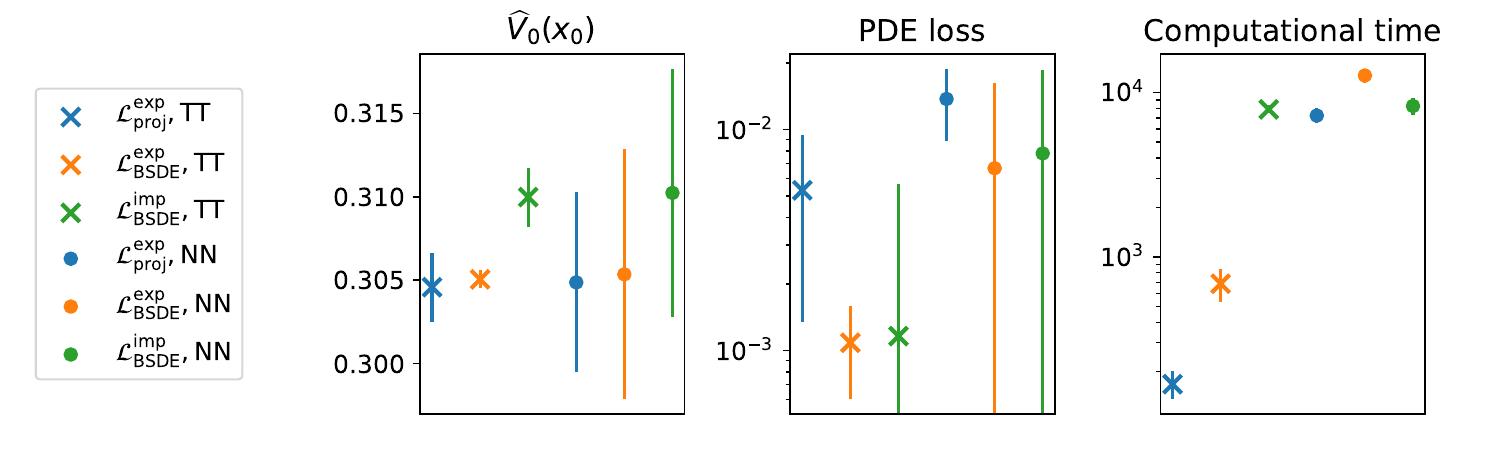}}
\caption{We compare different loss functions for a $100$ dimensional CIR example, either relying on tensor trains or neural networks. For computational details of the evaluation we refer to \Cref{app: implementation details}.}
\label{fig: CIR metrics}
\end{center}
\vskip -0.2in
\end{figure}

Finally, we move on to an example from financial mathematics. As proposed in \cite{jiang2021convergence}, we consider the bond price in a multidimensional Cox-Ingersoll-Ross model (CIR), see also \cite{hyndman2007forward} and \cite{alfonsi2015affine}. The PDE is of the form
\begin{equation}
    \partial_t V + \frac 1 2 \sum_{i,j = 1}^d \sqrt{ x_i x_j} \gamma_i \gamma_j \partial_{x_i} \partial_{x_j} V 
    + \sum_{i = 1}^d a_i (b_i - x_i) \partial_{x_i} V  - \left(\max_{1 \leq i \leq d} x_i \right) V = 0,
\end{equation}
where notationally we omit the space and time dependence of $V=V(x,t)$ for brevity. Here, $a_i, b_i, \gamma_i \in [0, 1]$ are uniformly sampled at the beginning of the experiment and the terminal condition is set to $V(T, x) = 1$. We set the dimension to $d = 100$ and for the tensor trains use polynomial degree $3$ and rank $1$. Since no reference solution is available we restrict ourselves to the PDE loss as a measure of accuracy.  We display the results in \Cref{fig: CIR metrics}, showing clear advantages of tensor trains over neural networks as well as the BSDE losses over the projection loss, noting that the explicit BSDE loss is much faster than the implicit one.

\section{Conclusions and outlook}
\label{sec: conclusion}

In this paper, building upon the work of \citet{richter2021solving}, we have demonstrated the efficacy of tensor trains as a compelling approximation framework for parabolic PDEs. Through a reformulation of the problem in terms of backward stochastic differential equations (\mbox{BSDEs}), we leverage algorithms that draw on backward-in-time iterations to efficiently solve for the PDE along simulated paths of a stochastic process.

Placing emphasis on continuous-time formulations, we have discussed three different loss functionals, one of which is  combines statistical robustness with efficient and fast algorithmic computations that has so far attracted very little interest from the community. 
Future work might consider applying tensor trains to elliptic PDEs and PDEs on bounded domains, and exploring the use of tensor trains with different algorithms such as variational formulations or residual minimization in the spirit of PINNs \citep{karniadakis2021physics}.

\acks This research has been partially funded by Deutsche Forschungsgemeinschaft (DFG) through the grant CRC $1114$ ``Scaling Cascades in Complex Systems'' (projects A$02$ and A$05$, project number $235221301$) and the Research Training Group ``Differential Equation- and Data-driven Models in Life Sciences and Fluid Dynamics: An Interdisciplinary Research Training Group (DAEDALUS)'' (GRK 2433).

\newpage

\appendix

\section{Proofs}
\label{app:proofs}

\begin{proof} \hspace{-0.1cm}\textbf{of \Cref{prop:variance0} (and Remark \ref{rem:vargradproj}).}
Let us define the shorthand
\begin{equation}
    h_s^{\varphi, (k)} := h(X_s^{(k)}, s, Y_s^{{\varphi},(k)}, Z_s^{{\varphi},(k)}) = h(X_s^{(k)}, s, \varphi(X_s^{(k)}, s), \sigma^\top \nabla \varphi(X_s^{(k)}, s)).
\end{equation}
For $\mathcal{L}^{(K)}_\text{BSDE}$ in \eqref{eq:Var0BSDE} we compute
\begin{align}
\begin{split}
    &\frac{\mathrm d}{\mathrm d \varepsilon}\Big|_{\varepsilon=0} {\mathcal{L}}_\text{BSDE}^{(K)}(\varphi + \varepsilon \psi) =\\ &\qquad\frac{2}{K}\sum_{k=1}^K \Bigg(\varphi(X_{t_n}^{(k)}, t_n) - V(X_{t_{n+1}}^{(k)}, t_{n+1}) - \int_{t_n}^{t_{n+1}} h_s^{\varphi,(k)} \mathrm ds +  \int_{t_n}^{t_{n+1}} (\sigma^\top \nabla \varphi)(X_s, s)\cdot \mathrm dW_s \Bigg) \\
    &\qquad\quad \Bigg(\psi(X_{t_n}^{(k)}, t_n)  - \int_{t_n}^{t_{n+1}} (\sigma^\top \nabla \psi)(X_s, s)\mathrm ds   \\    &\qquad\qquad-\int_{t_n}^{t_{n+1}}\Big( \partial_y h_s^{\varphi,(k)}\phi(X_{s}^{(k)}, s) + \nabla_z h_s^{\varphi,(k)}\cdot (\sigma^\top \nabla \psi)(X_{s}^{(k)}, s) \Big)\mathrm ds \Bigg).
\end{split}
\end{align}
Setting $\varphi = V$, we see that 
\begin{equation}
\nonumber
V(X_{t_n}^{(k)}, t_n) - V(X_{t_{n+1}}^{(k)}, t_{n+1}) - \int_{t_n}^{t_{n+1}} h_s^{\varphi,(k)} \mathrm ds +  \int_{t_n}^{t_{n+1}} (\sigma^\top \nabla \varphi)(X_s, s)\cdot \mathrm dW_s = 0,
\end{equation}
almost surely, for all $k=1,\ldots, K$, due to \eqref{eq: Ito t_n t_n+1}, implying \eqref{eq:Var0BSDE}.

To justify Remark \ref{rem:vargradproj}, we compute
\begin{align}
\begin{split}
\frac{\mathrm d}{\mathrm d \varepsilon}\Big|_{\varepsilon=0}& {\mathcal{L}}_\text{proj}^{(K)}(\varphi + \varepsilon \psi) =\\ 
&\frac{2}{K}\sum_{k=1}^K \left(\varphi(X_{t_n}^{(k)}, t_n) - V(X_{t_{n+1}}^{(k)}, t_{n+1}) - \int_{t_n}^{t_{n+1}} h_s^{V,(k)} \mathrm ds \right) \psi(X_{t_n}^{(k)}, t_n).
\end{split}
\end{align}
Setting $\varphi = V$ yields
\begin{align}
\label{eq:grad L proj}
\frac{\delta}{\delta \varphi}\Bigg|_{\varphi = V}\mathcal{L}_\mathrm{proj}^{(K)}(\varphi; \psi) = -\frac{2}{K}\sum_{k=1}^K \left(\int_{t_n}^{t_{n+1}} (\sigma^\top \nabla V)(X_s^{(k)}, s)) \cdot \mathrm dW_s^{(k)} \right) \psi(X_{t_n}^{(k)}, t_n).
\end{align}
According to the law of total variance, the variance of \eqref{eq:grad L proj} is given by
\begin{subequations}
\begin{align}
\label{eq:convar1}
\frac{4}{K^2}  \mathbb{E} \left[ \Var \left( \sum_{k=1}^K \left(\int_{t_n}^{t_{n+1}} (\sigma^\top \nabla V)(X_s^{(k)}, s)) \cdot \mathrm dW_s^{(k)} \right) \psi(X_{t_n}^{(k)}, t_n)\right) \Bigg\vert \mathcal{F}_n \right] 
\\
\label{eq:convar2}
+ \frac{4}{K^2} \Var \left(\mathbb{E} \left[ \sum_{k=1}^K \left(\int_{t_n}^{t_{n+1}} (\sigma^\top \nabla V)(X_s^{(k)}, s)) \cdot \mathrm dW_s^{(k)} \right) \psi(X_{t_n}^{(k)}, t_n) \right] \Bigg \vert \mathcal{F}_n \right). 
\end{align}
\end{subequations}
Since $\psi(X_{t_n}^{(k)},t_n)$ is $\mathcal{F}_n$-measurable, the contribution in \eqref{eq:convar2} vanishes according to the martingale property of the It{\^o} stochastic integral. Similarly, the term in \eqref{eq:convar1} equals \eqref{eq: variance of grad L_proj} by It{\^o}'s isometry.
\end{proof}

\section{Implementation details}
\label{app: implementation details}

In order to evaluate our approximations for solving the PDE in \eqref{eq: definition general PDE} we depend on reference values $V_\mathrm{ref}$ of $V$ at $(x, t) = (0,x_0)$. This allows us to calculate the \emph{relative error} of $\widehat{V}_0$ using
\begin{equation}
\mathcal{E}_\mathrm{rel} = \left|\frac{\widehat{V}_0(x_0) - V_\mathrm{ref}(x_0, 0)}{V_\mathrm{ref}(x_0, 0)}\right|.
\end{equation}
Following the computation of different approximations $(\widehat{V}_0^{(m)})_{1\le m\le M}$ in $M$ runs, we determine the \emph{root mean squared error (RMSE)}  by
\begin{equation}
\mathcal{E}_\mathrm{RMSE} = \sqrt{\frac{1}{M}\sum_{m=1}^M \left( \widehat{V}_0^{(m)}(x_0) - V_\mathrm{ref}(x_0, 0) \right)^2 }.
\end{equation}
Additionally, we introduce two error metrics that are (at least approximately) zero if and only if the PDE is satisfied along the samples generated by the discrete forward SDE in \eqref{eq: Euler forward SDE}.

First, we define the \textit{PDE loss} (inspired by \citet{raissi2019physics}) as
\begin{align*}
&\mathcal{E}_\mathrm{PDE} = \frac{1}{K N}\sum_{n=1}^{N} \sum_{k=1}^K\Big( (\partial_t + L) V(\widehat{X}_n^{(k)}, t_n) + h(\widehat{X}_n^{(k)}, t_n, V(\widehat{X}_n^{(k)}, t_n), (\sigma^\top \nabla V)(\widehat{X}_n^{(k)}, t_n))\Big)^2,
\end{align*}
where $\widehat{X}_n^{(k)}$ are realizations of \eqref{eq: Euler forward SDE}, the time derivative is approximated with finite differences and the space derivatives are computed analytically (or using automatic differentiation). 
We exclude the initial time step ($n=0$) due to the ill-defined regression problem ($X_0^k = x_0$ has the same value for all $k$) in the explicit and implicit tensor train schemes, but still achieve a good approximation as the regularization term provides a minimum norm solution with the correct point value $V(x_0,0)$.

Second, we establish the \textit{relative reference loss} as
\begin{align}
\mathcal{E}_\mathrm{ref} = \frac{1}{K (N+1)}\sum_{n=0}^{N} \sum_{k=1}^K\left| \frac{V(\widehat{X}_n^{(k)}, t_n)  - V_{\text{ref}}(\widehat{X}_n^{(k)}, t_n)}{V_{\text{ref}}(\widehat{X}_n^{(k)}, t_n)} \right|,
\end{align}
whenever a reference solution for all $x$ and $t$ is available.\par\bigskip

All computation times in the reported tables are measured in seconds.\par\bigskip

Our experiments have been performed on a desktop computer containing an AMD Ryzen Threadripper $2990$ WX $32$x $3.00$ GHz mainboard and an NVIDIA Titan RTX GPU, where we note that only the NN optimizations were run on this GPU, since our TT framework does not include GPU support. It is expected that running the TT approximations on a GPU will improve time performances in the future, see \cite{abdelfattah2016high}.\par\bigskip

All our code is available under \url{https://github.com/lorenzrichter/PDE-backward-solver}.\par\bigskip

For further details, such as tensor train settings and neural network architectures, we refer to Appendix D in \cite{richter2021solving}.

\vskip 0.2in
\bibliography{main}

\end{document}